\documentclass[letterpaper, 10 pt, conference]{ieeeconf}

\IEEEoverridecommandlockouts   
\usepackage{times}

\usepackage{graphicx}
\usepackage{subfig}
\usepackage{amsmath}
\usepackage{amssymb}
\usepackage{booktabs}
\usepackage{gensymb}
\usepackage{mathtools}
\usepackage{subcaption}
\usepackage{float}
\usepackage{array}

\newcommand{\chk}{\checkmark}
\newcommand{\X}{$\times$}

\begin{document}

\title{SpectralZoom: Efficient Segmentation with an \\ Adaptive Hyperspectral Camera}

\author{Jackson Arnold, Sophia Rossi, Chloe Petrosino, Ethan Mitchell\thanks{Authors are with the Department of Electrical and Computer Engineering, University of FLorida, Gainesville, FL, USA. Email: \{jarnold2, sophia.rossi, chloe.petrosino, mitchell.ethan, sjkoppal\}@ufl.edu}, 
and Sanjeev J. Koppal\thanks{Sanjeev J. Koppal holds concurrent appointments as an Associate Professor of ECE at the University of Florida and as an Amazon Scholar. This paper describes work performed at the University of Florida and is not associated with Amazon.}
}



%

\maketitle

\begin{abstract}
  Hyperspectral image segmentation is crucial for many fields such as agriculture, remote sensing, biomedical imaging, battlefield sensing and astronomy. However, the challenge of hyper and multi spectral imaging is its large data footprint. We propose both a novel camera design and a vision transformer-based (ViT) algorithm that alleviate both the captured data footprint and the computational load for hyperspectral segmentation. Our camera is able to adaptively sample image regions or patches at different resolutions, instead of capturing the entire hyperspectral cube at one high resolution. Our segmentation algorithm works in concert with the camera, applying ViT-based segmentation only to adaptively selected patches. We show results both in simulation and on a real hardware platform demonstrating both accurate segmentation results and reduced computational burden. 
\end{abstract}    
\section{Introduction}\label{sec:introduction}

Hyperspectral imaging (HSI)\footnote[3]{Defined here as images with more than three spectral channels.}  is the measurement of light both beyond and in-between the traditional red, green, and blue wavelengths perceived by the human eye and captured in a typical camera. Hyperspectral data is crucial in a variety of computer vision tasks, such as remote sensing, smart agriculture, astronomy and in the bio-sciences. In computer vision, HSI imaging is an ongoing topic with efforts in representation, invariant discovery, semantic segmentation and unmixing~\cite{healey1999invariant,zhang2020deep,yorimoto2021hypermixnet,janiczek2020differentiable,zhang2021learning}.

However, most current approaches involve capturing a large amount of HSI data from many different views. This creates post-processing related storage, transmission, and computing problems. For example, a drone flying over a 20 acre farm with a megapixel imager with a 100 channels can easily collect gigabytes of HSI data. Processing this data stream becomes additionally difficult when one notes that many HSI applications require multiple sets of imagery collected over time, resulting in repeated capture of data that only slightly differs from previous sessions.

\begin{figure}[t]
  \centering
  \includegraphics[width=1\linewidth]{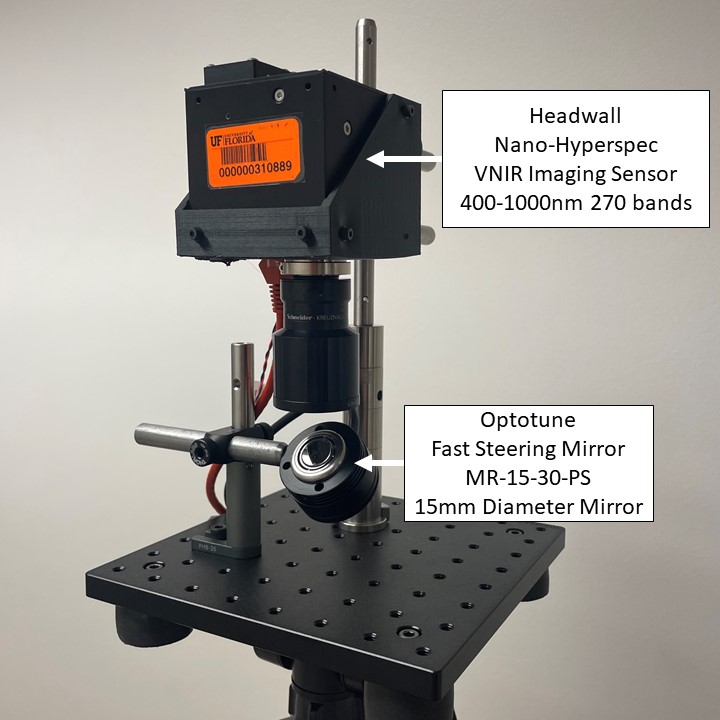}
  \caption{Hardware Prototype}
  \label{fig:proto}
\end{figure}

{\begin{table*}[h!]
\centering
\resizebox{\linewidth}{!}{
\begin{tabular}{|c|c|c|c|c|}
\hline                    
Method (with few examples) & Adaptive Patches & Hyperspectral & Attention \emph{during} image capture & Attention \emph{after} image capture  \\
\hline                  
Deep Attention Mechanisms \cite{vaswani2017attention,wang2020object,johnston2020self} & \X   & \chk & \X & \chk  \\
Learned Zoom \cite{zhang2019zoom} & \X & \X & \X & \X  \\
Adaptive Zoom \cite{Uzkent_2020_CVPR} & \chk  & \X & \X & \X  \\
SaccadeCam \cite{tilmon2021saccadecam} & \chk & \X & \chk  & \X  \\ 
\hline
\hline
SpectralZoom (\textbf{Ours}) & \chk & \chk & \chk & \chk  \\ 
\hline
\end{tabular}
}
\caption{SpectralZoom vs. Other Alternatives: To our knowledge, ours is the only work that provides hyperspectral segmentation through adaptive attention both during capture and after capture (via ViTs), along with the use of adaptive patches to reduce computation.} \label{fig:tabqual}
\end{table*}
}

In this paper, we demonstrate two solutions to the problem of too much HSI data. First, we create a camera design that adaptively controls the HSI resolution, up to the maximum angular resolution given by its diffraction limited lens. In other words, we can break the scene into patches of different sizes and resolution. Second, we note that this ``patch camera" can be integrated with vision transformers (ViTs) whose recent successes \cite{vaswani_attention_2017} have to do with self-attention of localized image patches during training. 

Our workflow also estimates an attention mask of the scene, which depicts which part of the scene should be captured at high resolution. Given an attention mask, our camera can capture much less data than naive equiangular sampling of the scene.

We first show experiments with a fixed attention policy where, for a given HSI task, the attention mask needed to maximize task accuracy is known. We also extend this to learned attention maps that are created in concert with the ViT system that takes in as input multi-resolution patches. Most of the patches are from a low-resolution ``background" image, while a small set are high-resolution patches in learned regions of interest. Compared to a ViT system that processes all the high-resolution patches, our approach reduces compute without sacrificing performance.  

Our driving application in this paper is hyperspectral segmentation, which is a crucial use of HSI data in identifying vegetation, water and soil in agricultural scenes, separating organisms from background in biological tasks and identifying cosmological objects in astronomy. We show efficient segmentation results both in simulation, on hyperspectral datasets, and on a real hardware platform demonstrating the tradeoff between accurate segmentation and reduced processing.  \textbf{Our contributions are:}


\begin{itemize}
    \item We created a hyperspectral camera-mirror prototype that can rapidly change the scene and field of view to capture wide angle images with a pushbroom hyperspectral camera as well as allow for super high resolution images after mosaicking.
    \item We modified a ViT transformer-based semantic segmentation algorithm for hyperspectral data. The transformer uses a small number of high-resolution patches suggested by the attention map, as well as low-resolution patches from the low resolution image, to infer scene segmentation. 
    \item We demonstrate our entire setup with simulated hardware experiments, showing that our system can adaptively capture patches and reduce both storage space, imaging time and compute load. 
\end{itemize}

\section{Related Work}
\label{sec:related}

In Table \ref{fig:tabqual} we compare our work to previous adaptive efforts. Our work is related to HSI imaging, ViT-based classification and other efforts that we describe below:

\textbf{Hyperspectral Imaging:} HSI has many impacts in agriculture, remote sensing, bio-sciences and astronomy. Classification is a major application due to additional color band information~\cite{ahmad_fast_2022, hu_lightweight_2021, chen_deep_2014, chen_hyperspectral_2011, el_rahman_hyperspectral_2016} Specifically, in the field of agriculture, HSI has allowed researchers to catalogue crops along stages of maturity and classify healthy versus stressed crops. \cite{gao_real-time_2020, varga_measuring_2021, zou_peanut_2019, nguyen_early_2021, aredo_predicting_2019, behmann_detection_2014}. In this work, we seek to enable such applications with a camera-algorithm co-design that reduces capture, storage and compute.

\textbf{HSI Cameras and Mirrors:} Push broom or line scan cameras are the primary type of hyperspectral cameras. These allow UAVs and satellites to image as they fly over various scenes~\cite{spoto_overview_2012,ortega_hyperspectral_2019}. We use a movable mirror with a pushbroom HSI camera, and we account for the rotational path of the virtual reflected view and the warping of the image after reflection \cite{somers_maintaining_2005}.
Many commercial HSI cameras also exist with mirror-based view modulation (e.g. Specim IQ, GaiaField, etc.), and research is ongoing to create low-cost versions of these~\cite{liu_fast_2020}. However, these systems offer only one scanning direction, while our system has two axis control, and further, we demonstrate mirror control in concert with a segmentation algorithm to increase the computational efficiency of processing HSI images.

\textbf{Vision Transformers:}
A relatively new architecture for image classification problems is the vision transformer proposed in \cite{dosovitskiy_image_2021} that is rapidly being applied to all manner of computer vision problems from semantic segmentation \cite{strudel_segmenter_2021} to denoising \cite{chen_pre-trained_nodate} with success. This new architecture relies on separating an image into patches and doing full attention maps in these patches \cite{vaswani_attention_2017}. Vision transformers are outperforming the current standard CNN structures, but they require more data and computations. Efforts have been made to reduce the data input by blotting out redundant patches \cite{pan_ia-red2_nodate} or reshaping patches \cite{tang_patch_2022}. Our SpectralZoom camera is uniquely designed to take advantage of these ``patch" based attention methods, and we have shown an adaptive version of a ViT segmenter that works by only selecting certain areas to image in high resolution. 

\textbf{Transformer based Spectral Classification:} Transformers have been applied to spectral imaging in \cite{hong_spectralformer_2021, yun_spectr_2021, zhang_adaptive_2020}. These have shown moderate success but exceptional need for training data and computational power. Because of the numerous color channels, both the spatial and spectral information has to be considered. The architecture in \cite{hong_spectralformer_2021} uses overlapping spectral data as the input to the transformer to solve some of the issues with the immense amount of spectral data and similar images.

\textbf{Adaptive Imaging and Vision:} Many adaptive cameras and algorithms have been proposed, including adaptive structured light patterns \cite{baek2020polka}, resolution for depth images~\cite{tilmon2021saccadecam},  eye-tracking~\cite{tilmon2020foveacam}, lens designs for depth estimation \cite{chang2019deep} and optimal bayer patterns \cite{chakrabarti2016learning}. What separates us from previous work is the co-design with a vision transformer for adaptive imaging in HSI.
In \cite{Uzkent_2020_CVPR}, a small, fixed number of high-resolution patches are selected to obtain better classification accuracy. Our approach is related to this work --- however, we both implement the patch selection on a real camera and we integrate the patch selection with a ViT system for efficient hyperspectral segmentation.

\section{Camera Model}

In Fig. \ref{fig:proto}, we show our prototype, which consists of a pushbroom hyperspectral camera whose line sensor is reflected off a galvo mirror. The mirror is situated at a 45\degree angle of incidence. We assume the camera before reflection is an instance of the linear pushbroom model \cite{gupta1997linear}. We also assume the entire line resolution fits onto the mirror, using the same approach as done by \cite{tilmon2020foveacam}.

\def\imghspace{\hspace{2pt}}
\captionsetup[subfigure]{labelformat=empty}
\begin{figure*}[t]
\centering

\subfloat[Uncorrected RGB]{\includegraphics[width=0.15\linewidth]{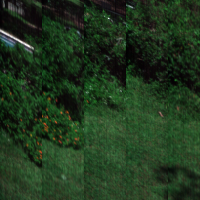}}\hspace{0.01\linewidth}
\subfloat[Corrected RGB]{\includegraphics[width=0.15\linewidth]{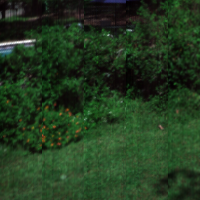}}\hspace{0.01\linewidth}
\subfloat[$\lambda$ = 450nm]{\includegraphics[width=0.15\linewidth]{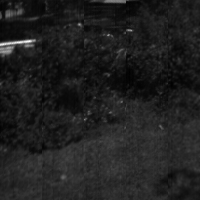}}\imghspace
\subfloat[$\lambda$ = 550nm]{\includegraphics[width=0.15\linewidth]{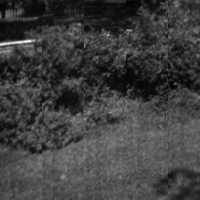}}\imghspace
\subfloat[$\lambda$ = 650nm]{\includegraphics[width=0.15\linewidth]{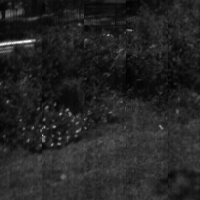}}\imghspace
\subfloat[$\lambda$ = 750nm]{\includegraphics[width=0.15\linewidth]{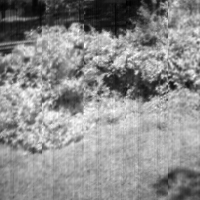}}

\caption{Raw imagery from our device: Here we show a few images captured outdoors using our mirror modulated linear pushbroom camera. The raw images are before our calibration and counter rotation is applied. After processing, the corrected images are the inputs into our wandering patch network.}
\label{fig:pics}
\end{figure*}

\subsection{Mirror Control}

The position of the mirror can be controlled by sending static position values between -1 and 1 to either of the axes. These values are referred to as $XY$ position values. The relationship between a position value $X$ and the angle of rotation $\theta$ is


\begin{equation}
  X = \frac{\tan{\theta}}{\tan{\theta_{max}}}
  \label{eq:xy2ang}
\end{equation}

\noindent where $\theta_{max}$ is the maximum angular field-of-view (FOV) of the mirror, 50\degree. A position value vector can be stored in the mirror controller's memory, and the mirror will cycle through these in sequence. The maximum length of this vector for each axis is 1500 values. The sampling speed of this position vector can be configured before the sequence run is initiated. 

Since the mirror is positioned at a 45\degree angle, the relationship between its motion and the scene can be found using a rotation. Let $\boldsymbol{o}$ be the path of the desired viewing direction, given by incident beam from the scene, reflected off of the mirror. We show how to find the $XY$ values for $\boldsymbol{o}$ in a few steps. First,  

\begin{equation}
    \boldsymbol{o} = \boldsymbol{R} \boldsymbol{i}
\end{equation}

\noindent where $\boldsymbol{i}$ is the path of the incident beam from the camera, and $\boldsymbol{R}$ is the 3D reflection matrix built from the mirror normal $\boldsymbol{n}_{m1} = \begin{bsmallmatrix} a & b & c \end{bsmallmatrix} ^T$ using  

\begin{equation}
    \boldsymbol{R} = 
    \begin{bmatrix}
    1-2a^2 & -2ab & -2ac \\
    -2ab & 1-2b^2 & -2bc \\
    -2ac & -2bc & 1-2c^2
    \end{bmatrix}
\end{equation}

We want the entire resolution of the camera to be centered around the patch corresponding to the desired viewing direction $\boldsymbol{o}$. We do this by setting the incident vector from the camera as $\boldsymbol{i} = \begin{bsmallmatrix} 0 & 0 & 1 \end{bsmallmatrix}^T$, and the mirror normal can be found from the objective vector with 

\begin{equation}
    \begin{split}
        a = \frac{\boldsymbol{o}_x}{2c} \\
        b = \frac{\boldsymbol{o}_y}{2c} \\  
        c = \sqrt{\frac{\boldsymbol{o}_z + 1}{2}}
    \end{split}
\end{equation}

Once the mirror normal vector is found, it can be rotated 45\degree so that the undeformed mirror normal $\boldsymbol{n}_{m0}$ would be $\begin{bsmallmatrix} 0 & 0 & 1 \end{bsmallmatrix} ^T$. This is done with 

\begin{equation}
    \boldsymbol{n}_{m0} = 
    \begin{bmatrix}
    \cos{45\degree} & 0 & \sin{45\degree} \\
    0 & 1 & 0 \\
    -\sin{45\degree} & 0 & \cos{45\degree}
    \end{bmatrix}
    \times
    \boldsymbol{n}_{m1}
\end{equation}

And the final $XY$ position that moves the mirror normal to face the viewing direction can be found by 

\begin{equation}
    \begin{split}
        X = \frac{\boldsymbol{n}_{m0_x}}{\sin{(\theta_{max}/2)}} \\
        Y = \frac{\boldsymbol{n}_{m0_y}}{\sin{(\theta_{max}/2)}}   
    \end{split}
\end{equation}

By creating an array of such objective vectors $\boldsymbol{o}$ in sequence, the corresponding $XY$ values for the mirror's position value vector can be built and sent to the controller. 

\begin{figure*}[thb]
    \centerline{\includegraphics[width=\textwidth]{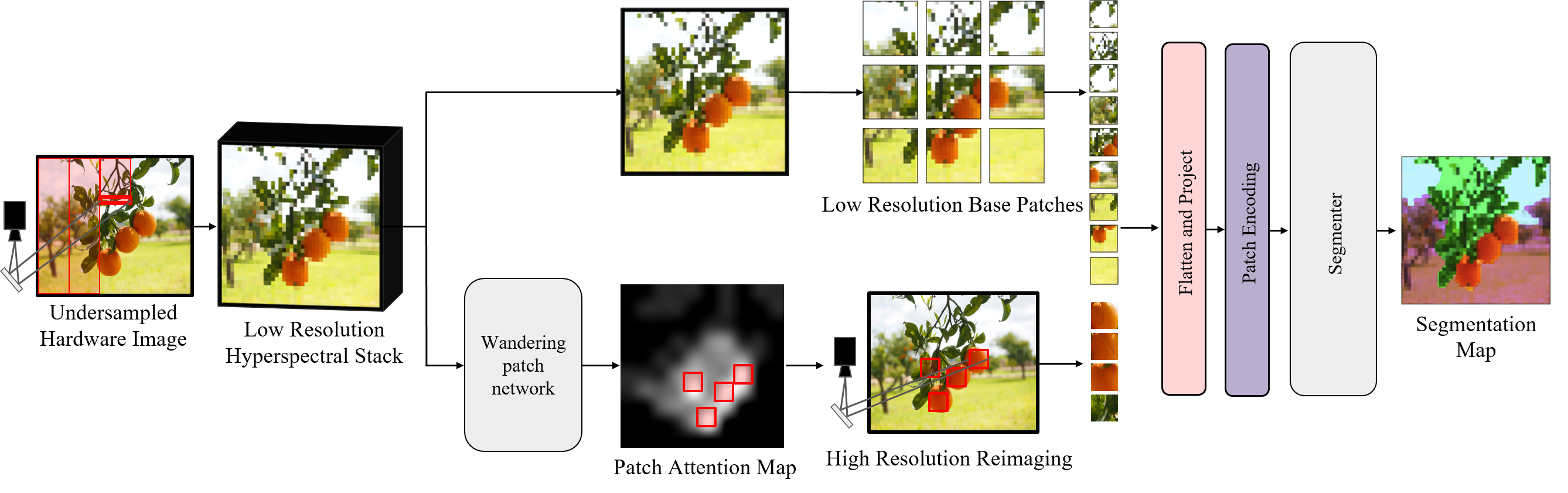}}
    \caption{Spectral zoom pipeline diagram. Our workflow is shown on the left, where the mirror modulates quickly across the field-of-view creating a low-resolution HSI image. This image is then sent to a attention network that produces an attention map. The camera then uses the attention map to select high-res patches. A small number of low-res patches from the low-res image plus high-res patches from the camera are combined into a ViT segmentor from \cite{strudel_segmenter_2021} that is modified to accept (a) HSI data and (b) variable patches. }
    \label{fig:block}
\end{figure*}

\begin{figure}[h]
  \centering
  \includegraphics[width=1\linewidth]{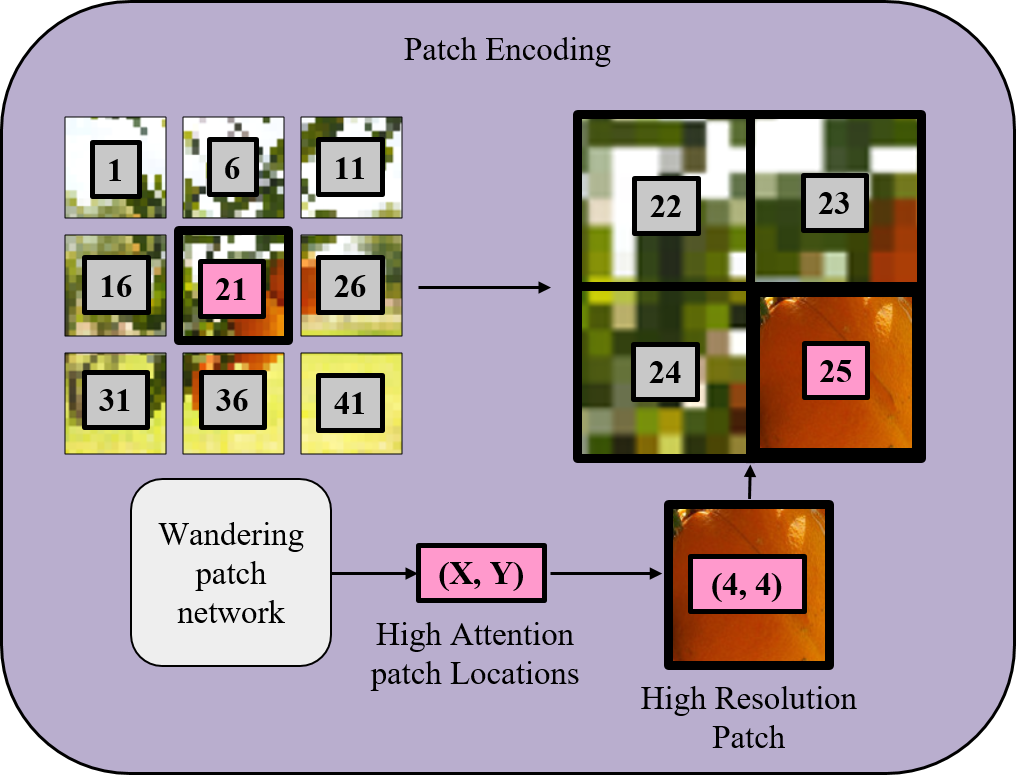}
  \caption{Wandering patch encoding scheme}
  \label{fig:patchencoding}
\end{figure}

\subsection{Mirror Sample Speed}

In a traditional push broom HSI setup, the scene would be moving below the camera or the camera itself would be moving with the aid of a drone or satellite. In our scenario, we discuss how to control the mirror to get acceptable images, i.e. where the captured pixels would 
be square and there would be no gaps between frames. 

In a conventional scenario, this would be achieved by calculating the speed $s_{Scene}$ of the drone or conveyor as a function of the exposure $t_{Exp}$, the distance to the target $d$, the sensor size $H$, and the focal length of the lens $f$. 

\begin{equation}
  s_{Scene} = \frac{H d}{f t_{Exp}} 
  \label{eq:drone speed}
\end{equation}

In our case, the speed of the scene is the speed of the mirror's movement, which is the sampling speed of the mirror $s_{sample}$. If the number of frames to be captured equals or exceeds the maximum mirror memory (1500 values), the mirror sampling speed will equal the frame period, and the image will be taken in steps where new mirror values will be saved to the mirror memory. In the case where the number of frames captured is less than the maximum mirror memory, the sample speed can be found by dividing the maximum mirror memory, $m_{Mem}$, by the time to image a scene in milliseconds, $t_{img}$. 

\begin{equation}
  s_{Sample} = \frac{m_{Mem}}{t_{Img}} 
  \label{eq:sample speed}
\end{equation}

The imaging time, $t_{Img}$, can be found from the capture time for each frame or the exposure $t_{Exp}$ multiplied by the number of frames to be captured or the y resolution of the image $y_{Res}$.

\begin{equation}
  t_{Img} = t_{Exp} y_{Res} 
  \label{eq:capture time}
\end{equation}

With control over the number of frames captured and the exposure time we can image a scene with perfect square pixels at full resolution. Additionally, We can increase the mirror speed to cover more of a scene, or increase the exposure to capture the same scene in fewer frames. The combination of these effects results in a lower resolution image of a larger scene captured in the same amount of time as the full resolution scene. 

\def\imghspace{\hspace{2pt}}

\begin{figure*}[thb]
\centering

    \subfloat[Input image]{\includegraphics[width=0.19\linewidth]{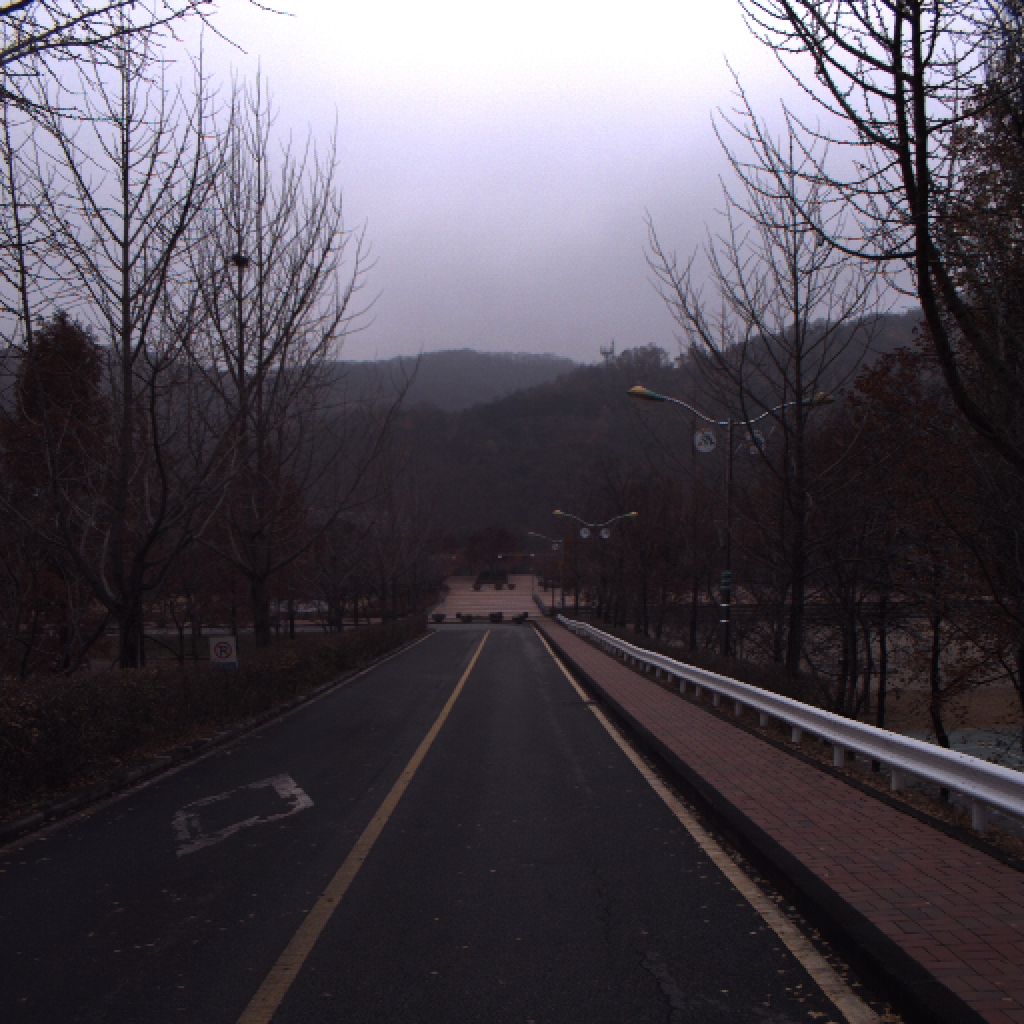}}\imghspace
    \subfloat[Sailency Map]{\includegraphics[width=0.19\linewidth]{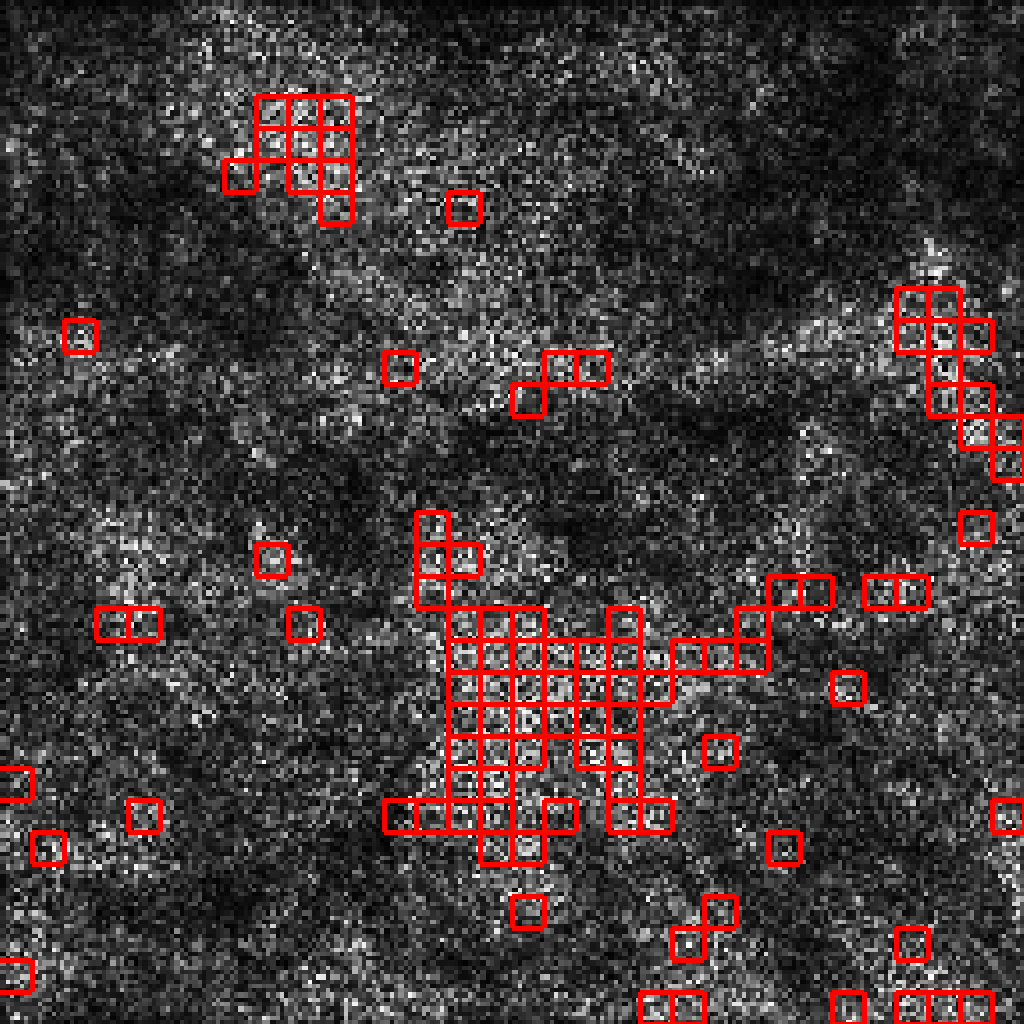}}\imghspace
    \subfloat[Ground Truth]{\includegraphics[width=0.19\linewidth]{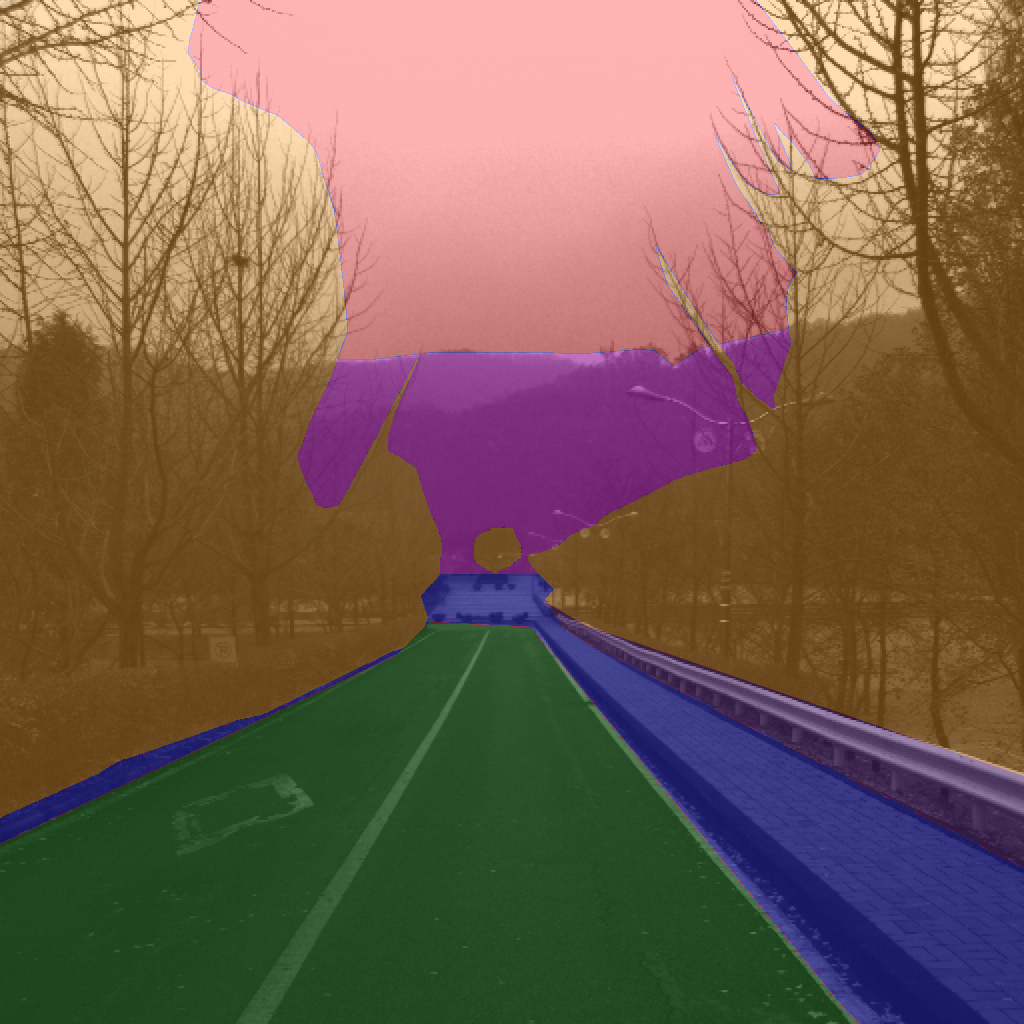}}\imghspace
    \subfloat[Segmentations]{\includegraphics[width=0.19\linewidth]{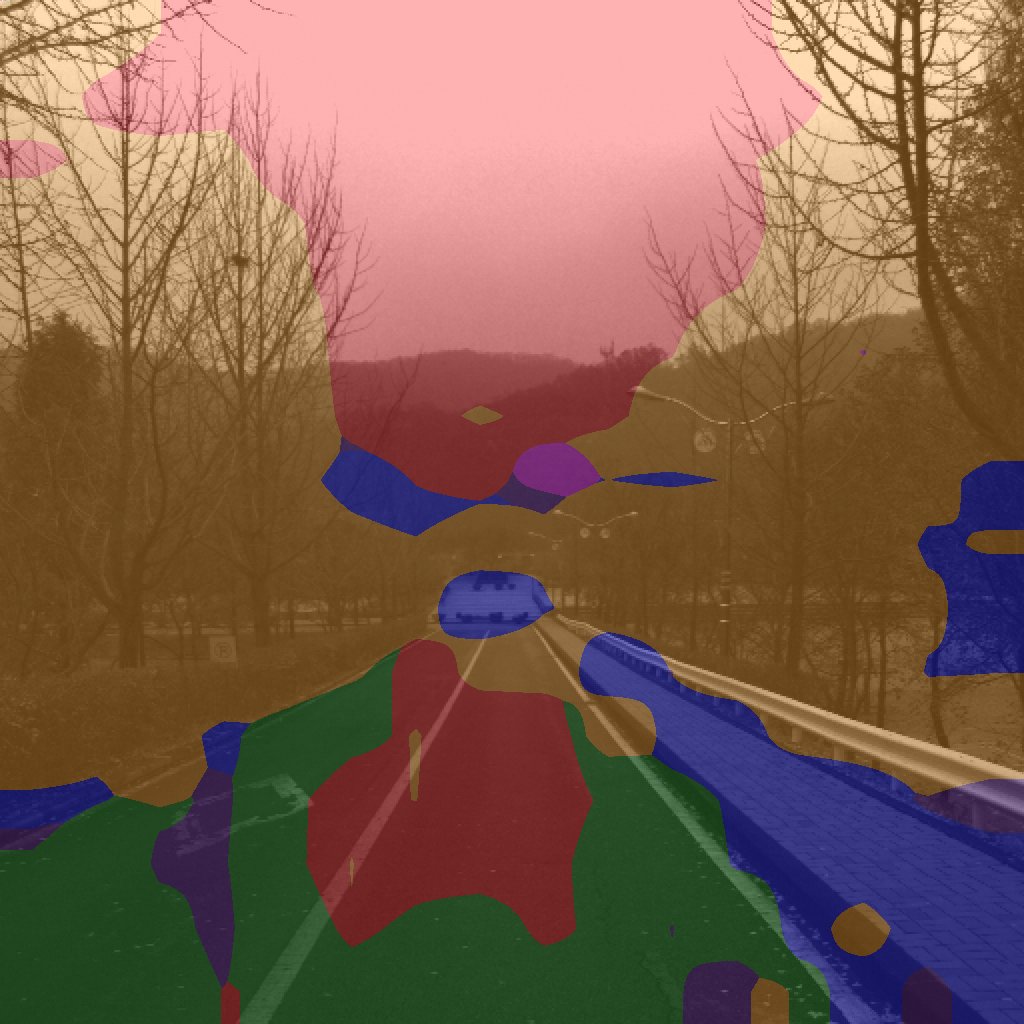}}\imghspace
    \subfloat[Baseline]{\includegraphics[width=0.19\linewidth]{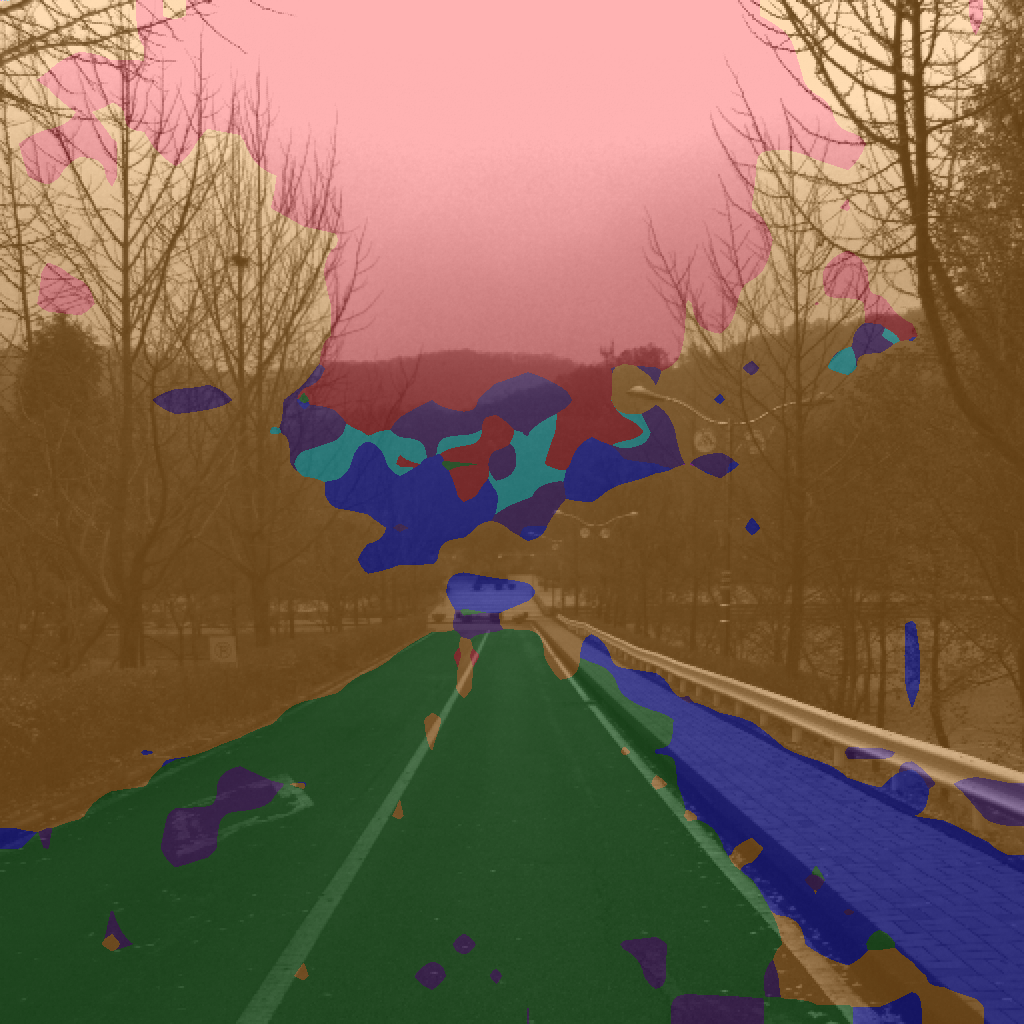}}
    \caption{512x512 input into the trainable sailency map with 100 wandering patches}


    \subfloat[Input image]{\includegraphics[width=0.19\linewidth]{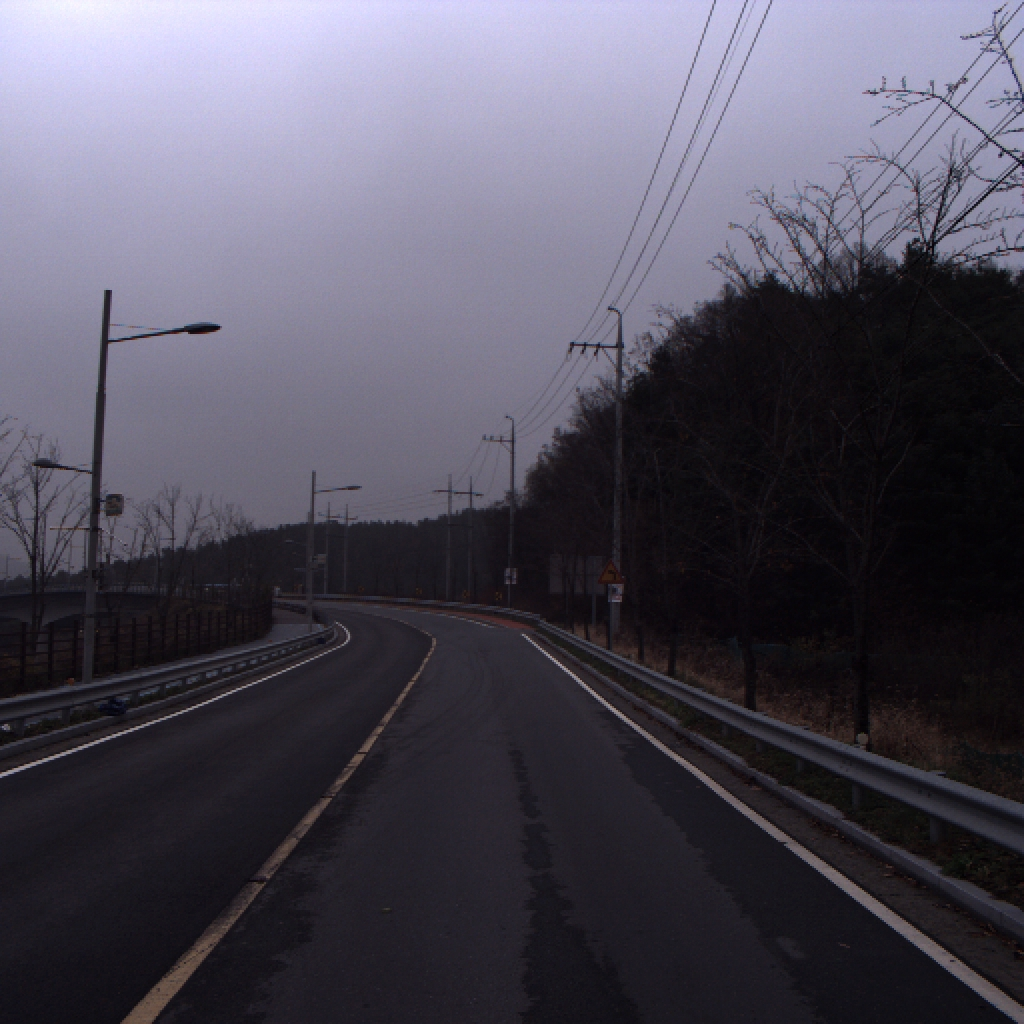}}\imghspace
    \subfloat[Sailency Map]{\includegraphics[width=0.19\linewidth]{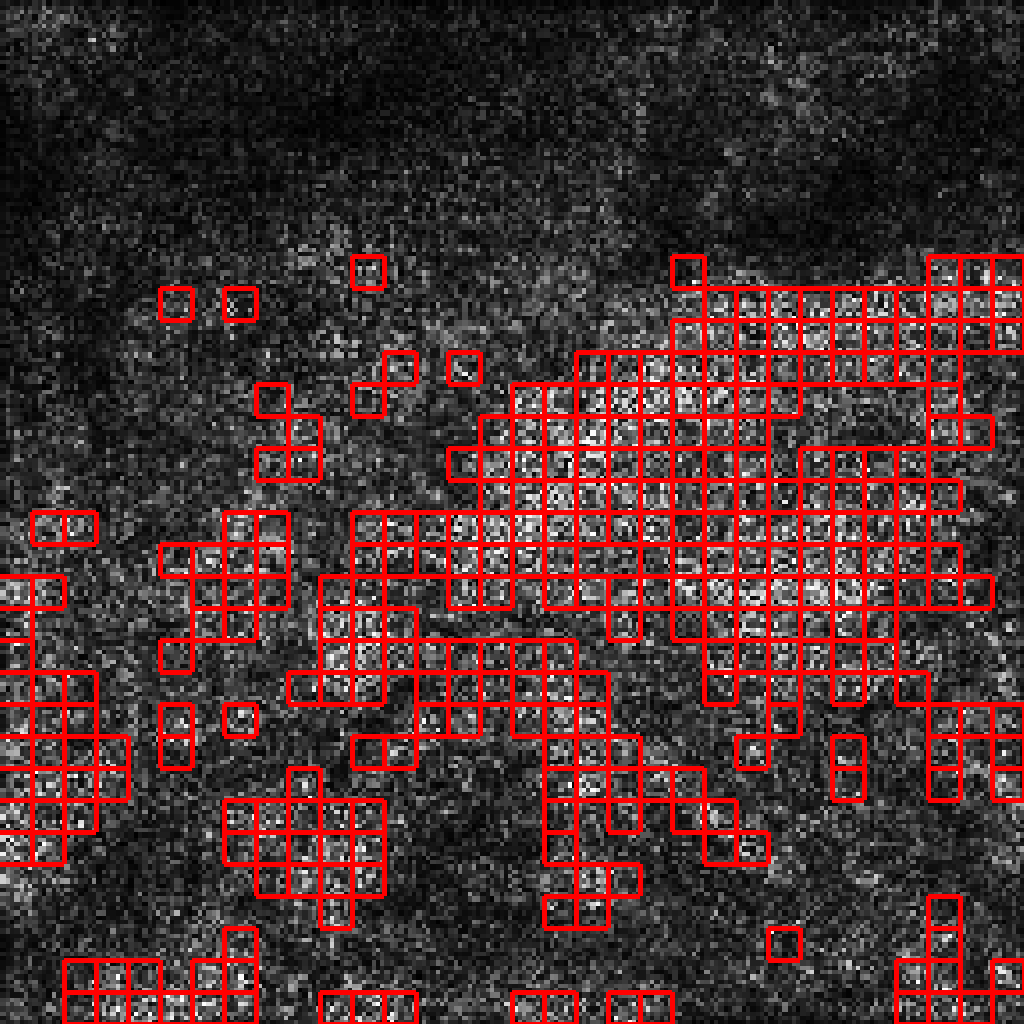}}\imghspace
    \subfloat[Ground Truth]{\includegraphics[width=0.19\linewidth]{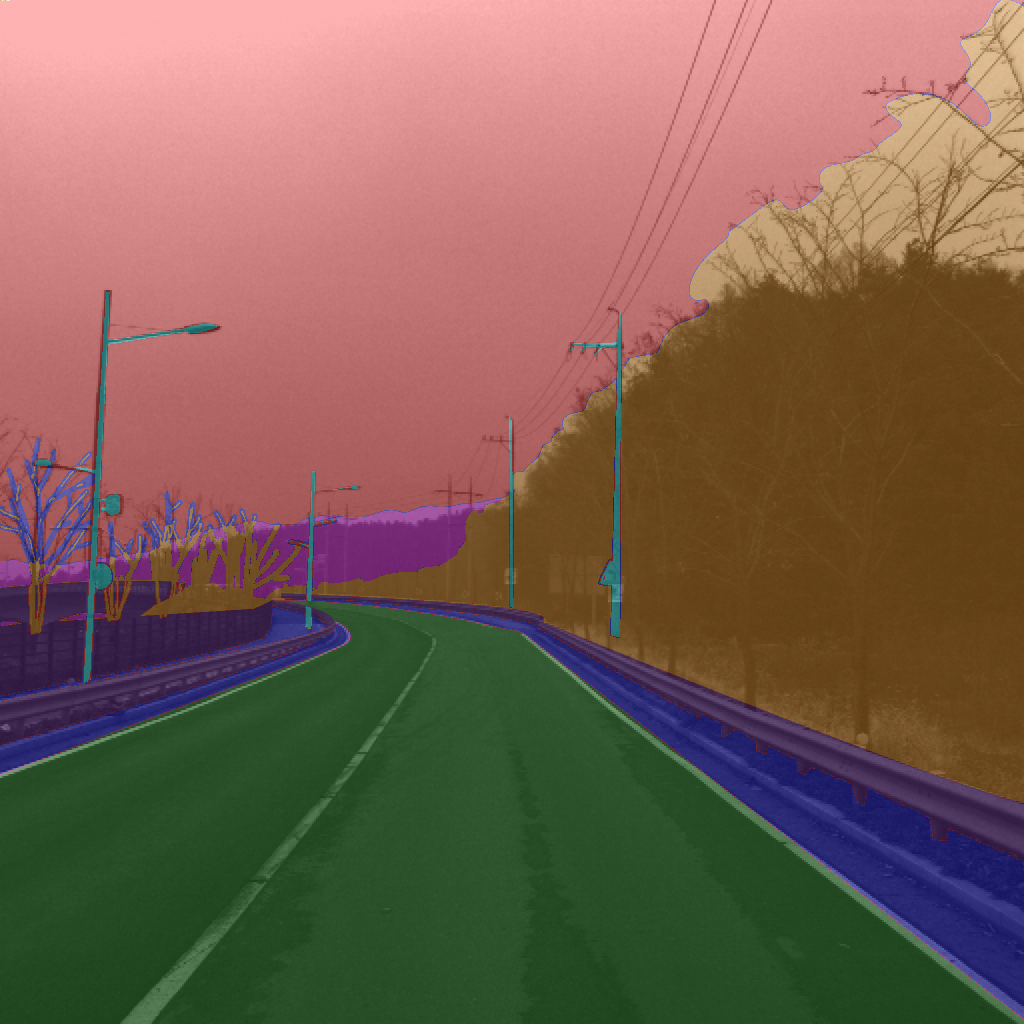}}\imghspace
    \subfloat[Segmentations]{\includegraphics[width=0.19\linewidth]{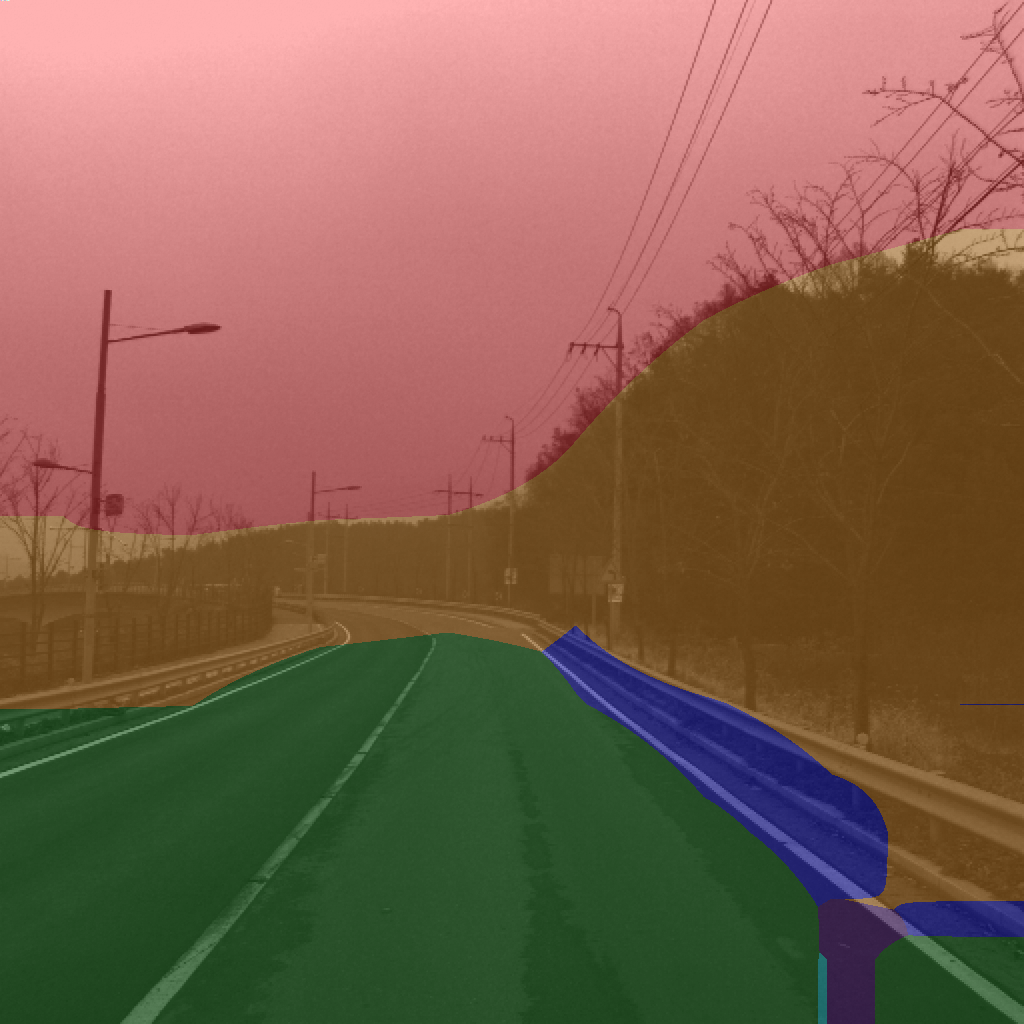}}\imghspace
    \subfloat[Baseline]{\includegraphics[width=0.19\linewidth]{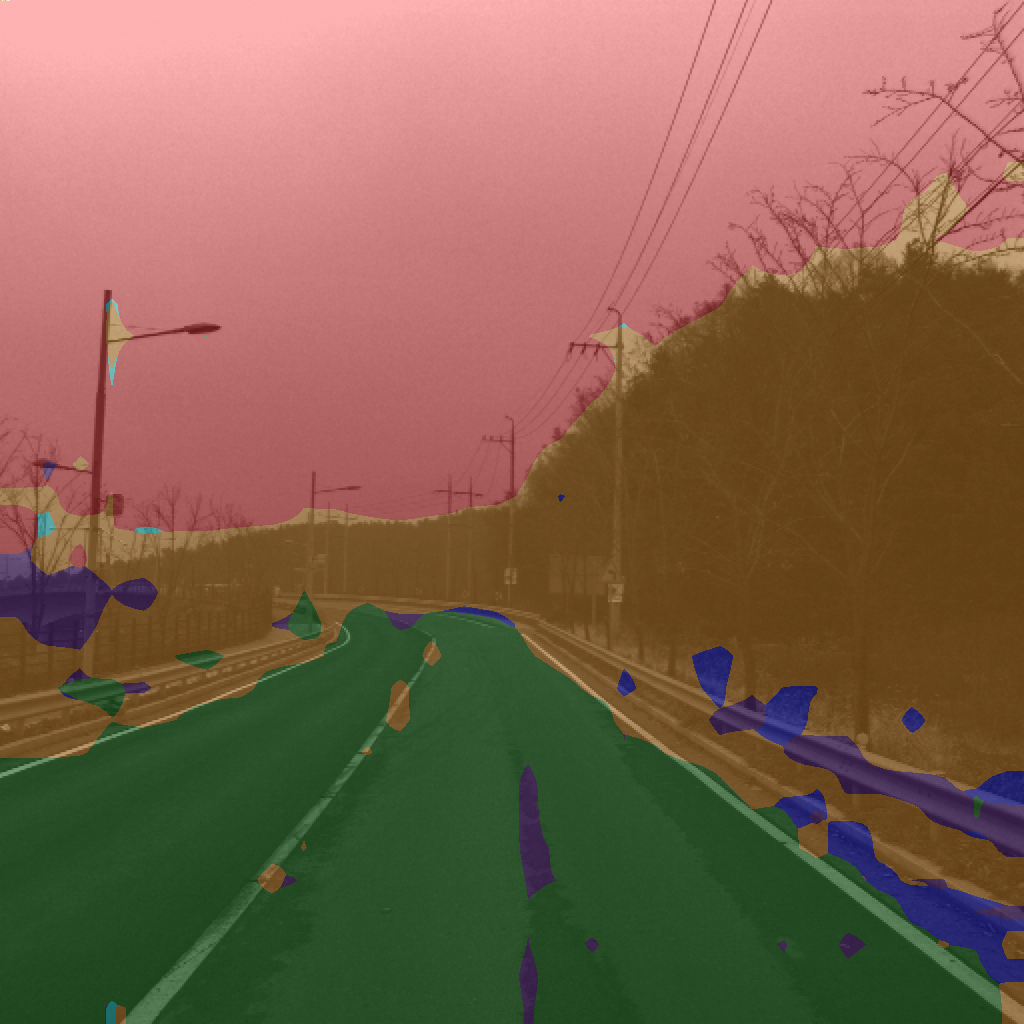}}    
    \caption{512x512 input into the static sailency map with 400 wandering patches}

\label{512 PA 100.fig}
\end{figure*}

\noindent \textbf{Prototype specifics:} In Fig. \ref{fig:proto}, we show our prototype with a Headwall Nano-Hyperspec hyperspectral camera. This is a linear pushbroom camera with a spatial resolution of 640p, a spectral resolution of 270 bands, a wavelength range of 400-1000nm, and a lens with focal length of 70mm. Images are collected using Headwall's Hyperspec III software and its associated python sdk. In conjunction with the camera, we use Optotune's 2 axis movable mirror, MRE-15-30. This mirror has a diameter of 15mm and is movable +/- 25\degree in both directions. 

\noindent \textbf{Imaging Workflow:} In Fig. \ref{fig:pics} we show some of the raw images captured from our system. By controlling the mirror sampling speed and the exposure time per frame, we can control the amount of information recorded per scene. For instance, a 640x600 image with aligned pixels and an exposure time of 100ms would take 60 seconds. If we were to disregard the aligned pixels and instead take a 640x300 image with 200ms of exposure, it would be captured in the same amount of time with half of the data collected. Or instead, if we were to double the distance covered in the reflected beam, but keep the sample speed the same, we would capture the same 640x600 image but effectively double the field of view in the same amount of time. In either of these processes, spatial resolution is sacrificed, but there are clear benefits in FOV and data storage.

\subsection{Image Alignment}
When an image is deflected from an angled mirror the image will be rotated. We needed to fix this in order to stitch our images together seamlessly. To correct this rotation we did the following: 
After deciding the desired camera projection path, the mirror coordinates for this path are calculated as per the methods described in the previous sections. The $XY$ values that determine this mirror path can be represented as a curve when $Y$ is plotted as a function of $X$. 

The tangent line along the path of a single image sweep corresponds to the rotation of each line of pixels. To find the counter rotation angle, we take the derivative of the $XY$ curve, then take the inverse tangent of the found slope. 

\begin{equation}
  \theta = -tan^{-1}(\mathrm{d}(XY_{curve}))
  \label{eq:sample speed}
\end{equation}

This gives us the rotation angle, and we counter rotate each frame with the corresponding angle. This process has no effect on the raw spectral values of the image.

\section{Multispectral Segmentation}

In this section we outline a novel 
deep segmentation algorithm that is designed to be used with our new camera. We have adapted this algorithm from a vision-transformer (ViT) segmenter for color images. We have also made changes to the architecture to support additional high resolution patches as inputs. 


\subsection{Multispectral Segmenter}
We have adapted an algorithm named Segmenter \cite{strudel_segmenter_2021} which performs semantic segmentation that is exclusively transformer based. Modifying an algorithm for multispectral data increased the number of channels throughout the network, increasing memory and computation. 

The original algorithm starts with a pretrained ViT core that contains weights after training on the ImageNet dataset. We also take advantage of this core, which jumpstarts the learning process. 

Outside this core, an additional self-attention mechanism processes the patches to produce semantic segmentation output. Since have modified this outer core to take additional channels, we cannot use the pretrained weights from \cite{strudel_segmenter_2021}, but instead must train from scratch on new hyperspectral data. 



\newcolumntype{P}[1]{>{\centering\arraybackslash}p{#1}}

\begin{table*}[h]
\centering
\begin{tabular}{|c|c|c|c||c|c|c|c|}
\hline
    \multicolumn{8}{|c|}{RANUS Dataset - 4 Color Channels} \\
 \hline
 \hline

  \multicolumn{4}{|c||}{Input image downsampled from 1024 to 256} & \multicolumn{4}{c|}{Input image downsampled from 1024 to 512}\\

  \hline
 
  & \multicolumn{1}{c|}{\centering Mean IOU} 
  & \multicolumn{1}{m{1.4cm}|}{\centering Pixel Accuracy} 
  & \multicolumn{1}{m{1.4cm}||}{\centering Total Patches} 
  &
  & \multicolumn{1}{c|}{\centering Mean IOU} 
  & \multicolumn{1}{m{1.4cm}|}{\centering Pixel Accuracy} 
  & \multicolumn{1}{m{1.4cm}|}{\centering Total Patches} \\
  \hline

  Baseline 256  & 22.62 & 68.62 & 64 & Baseline 512 & 24.14 & 69.47 & 256\\
  Baseline 1024 & 25.08 & 71.06 & 1024 & Baseline 1024 & 25.08 & 71.06 & 1024\\
  
  \hline
  ED 100 & 22.89 & 70.09 & 164 & ED 100 & 23.29 & 70.87 & 356\\
  \textbf{ED 200} & \textbf{24.80} & \textbf{70.79} & \textbf{264} & ED 200 & 21.26 & 69.60 & 456\\
  ED 300 & 22.58 & 68.18 & 364 & ED 300 & 21.73 & 69.26 & 556\\
  ED 400 & 21.78 & 67.69 & 464 & ED 400 & 23.65 & 71.70 & 656\\
  ED 500 & 22.16 & 67.54 & 564 & ED 500 & 24.37 & 72.69 & 756\\
  \hline
  SA 100 & 18.60 & 65.43 & 164 & SA 100 & 24.67 & 73.07 & 456\\
  SA 200 & 21.09 & 69.59 & 264 & SA 200 & 23.31 & 70.44 & 556\\
  SA 300 & 20.91 & 69.20 & 364 & SA 300 & 23.07 & 68.43 & 656\\
  SA 400 & 22.38 & 69.92 & 464 & SA 400 & 23.95 & 70.09 & 756\\
  SA 500 & 20.39 & 67.73 & 564 & SA 500 & 24.85 & 70.64 & 856\\
  \hline
  TA 100 & 23.31 & 70.90 & 164 & \textbf{TA 100} & \textbf{25.61} & \textbf{71.76} & \textbf{356} \\
  TA 200 & 22.99 & 70.21 & 264 & TA 200 & 24.75 & 71.80 & 456\\
  TA 300 & 23.51 & 70.71 & 364 & TA 300 & 25.32 & 72.35 & 556\\
  TA 400 & 22.35 & 69.40 & 464 & TA 400 & 25.27 & 71.57 & 656\\
  TA 500 & 23.08 & 69.06 & 564 & TA 500 & 24.28 & 71.04 & 756\\
  \hline
  
\end{tabular}
\caption{Comparison of each of the attention mechanisms to the baselines tested on a patch size of 32. Each section should be compared to its baseline. As the amount of patches increase, it should approach the results of the 1024 baseline, which would be if the image had access to all high resolution patches. The advantage to our model is that it does not need access to the entire high resolution image to achieve similar results. The model performed best when it was allowed to train its attention mechanism, and achieved a higher MIOU than its corresponding baseline.}
\label{table:segmentation results}
\end{table*} 

\begin{table}[!htb]
\centering
\begin{tabular}{|c|c|P{1.4cm}|P{1.4cm}|}
    \hline
    \multicolumn{4}{|c|}{HyKo Dataset - 15 Color Channels} \\
    \hline
    \hline
    & \multicolumn{1}{|c|}{\centering Mean IOU} 
    & \multicolumn{1}{m{1.4cm}|}{\centering Pixel Accuracy} 
    & \multicolumn{1}{m{1.4cm}|}{\centering Total Patches} \\
    \hline
    \hline
    Baseline 256  & 58.85 & 74.64 & 64 \\
    Baseline 1024 & 61.79 & 75.13 & 1024 \\
    \hline
    TA 100 & 53.12 & 67.63 & 164 \\
    TA 200 & 52.80 & 67.81 & 264 \\
    TA 300 & 52.87 & 68.19 & 364 \\
    TA 400 & 53.17 & 68.48 & 464 \\
    TA 500 & 54.66 & 69.41 & 564 \\
    \hline
    \hline
    Baseline 512  & 58.70 & 72.23 & 256 \\
    Baseline 1024 & 61.79 & 75.13 & 1024 \\
    \hline
    TA 100 & 58.29 & 72.06 & 356 \\
    TA 200 & 55.59 & 70.46 & 456 \\
    TA 300 & 58.62 & 72.50 & 556 \\
    TA 400 & 58.48 & 72.24 & 656 \\
    TA 500 & 59.06 & 72.68 & 756 \\         
    \hline
  
\end{tabular}
\caption{Comparisons on the HyKo hyperspectral dataset \cite{hyko}. The dataset consists of 15 color channels in the visible spectrum. All experiments were trained with the trainable sailency map attention mechanism as this is consistently the highest performing.}
\label{table:hyko table}
\end{table}

\subsection{Wandering Patch}

Unique to our algorithm are additional higher resolution patches that we term as wandering patches. The wandering patch is a higher spatial resolution image but the same number of pixels as the original patch inputs. These additional patches would provide the benefits of a high resolution image but localized to the areas where it is needed most such as high frequency regions of color change or edges. 

Selecting where to place the wandering patches is crucial to the performance of the overall network. In our setup, using the low-resolution image as input, we select these high-resolution patches with a network that we call the wandering patch network. 

This network will take the low resolution image as input and using the segmentation accuracy and the transformer's self-attention head attention map as the loss will output the ideal patch locations. The patch will then find a seat within one of the low resolution patches. 

\subsubsection{Wandering Patch Position Encoding}

Our trick to apply the position encoding is illustrated in Fig \ref{fig:patchencoding}. In this example, we split the patches in a quad-tree fashion. We encode the original patches in the low-resolution image with jumps of four. This allows any new wandering patch to be placed in the positional gaps in the encoding. For example, as in the figure, high resolution patch 25 is placed in one of the the gaps left between low resolution patches 21 and 26.

More generally, each low resolution patch can be divided into a number of smaller patches which we will refer to as the wandering patch seats. Each seat and each low resolution patch will need to have its own unique position encoding. The original low resolution image is separated into appropriate number of patches, $n$, each of the low resolution patches is divided into a number of potential wandering patch seats, $m$. The encoding for each low resolution patch is 

\begin{equation}
    P_{n} = n_i(m + 1)+1
\end{equation}

where $n_i$ is the respective low resolution patch. The encoding for the wandering patch seats is 

\begin{equation}
    P_{m} = P_{n} + m_i
\end{equation}

where $m_i$ is the respective seat position. This encoding method is shown in Fig. \ref{fig:patchencoding}. Using these positions, the patches are encoded according to the sine and cosine methods in \cite{vaswani_attention_2017}.

\def\imghspace{\hspace{2pt}}
\begin{figure*}[t]
\centering

\subfloat[Input image]{\includegraphics[width=0.16\linewidth]{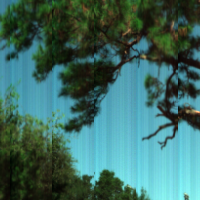}}\imghspace
\subfloat[Sailency Map]{\includegraphics[width=0.16\linewidth]{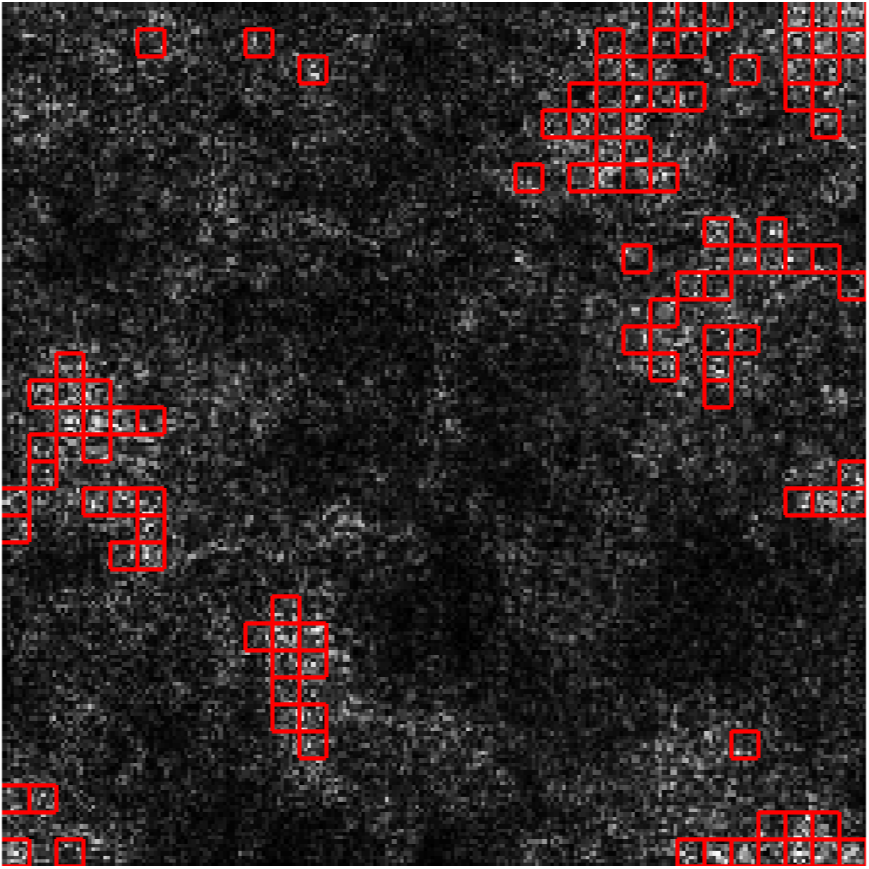}}\imghspace
\subfloat[Segmentations]{\includegraphics[width=0.16\linewidth]{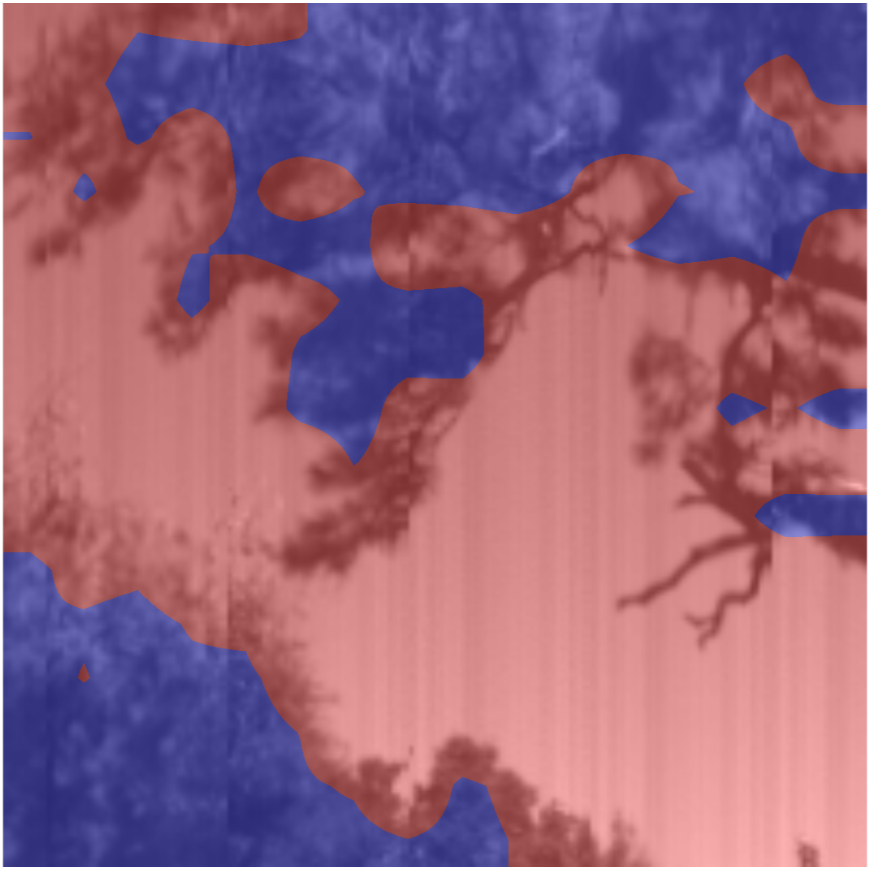}}\hspace{.01\linewidth}\setcounter{subfigure}{0}
\subfloat[Input image]{\includegraphics[width=0.16\linewidth]{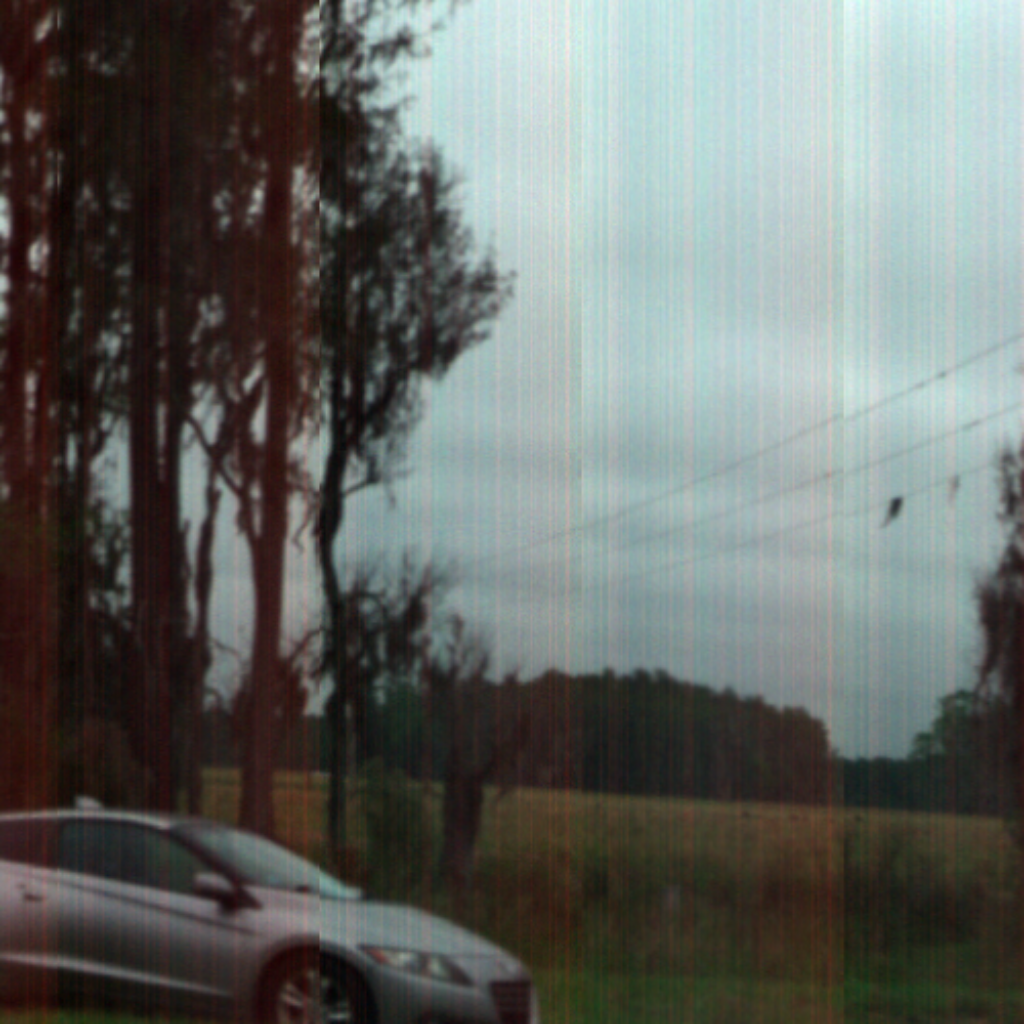}}\imghspace
\subfloat[Sailency Map]{\includegraphics[width=0.16\linewidth]{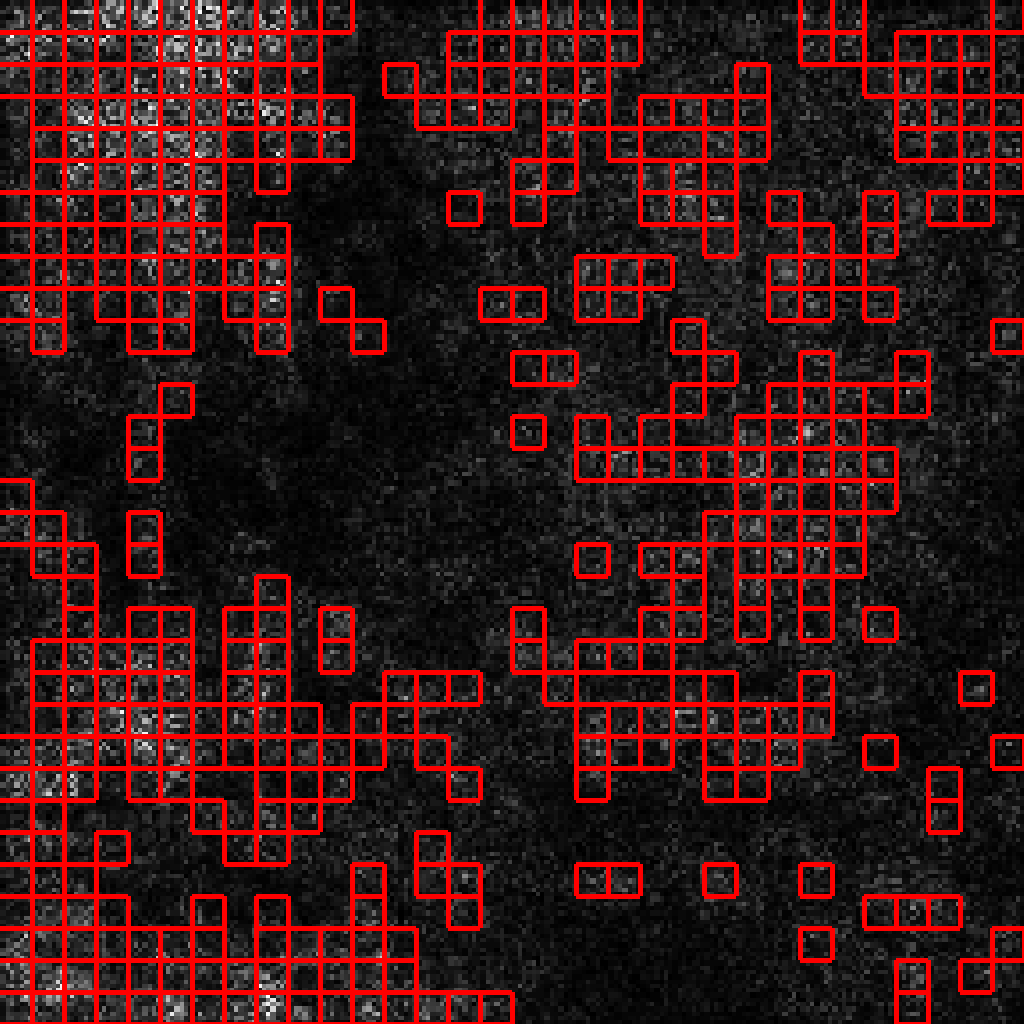}}\imghspace
\subfloat[Segmentations]{\includegraphics[width=0.16\linewidth]{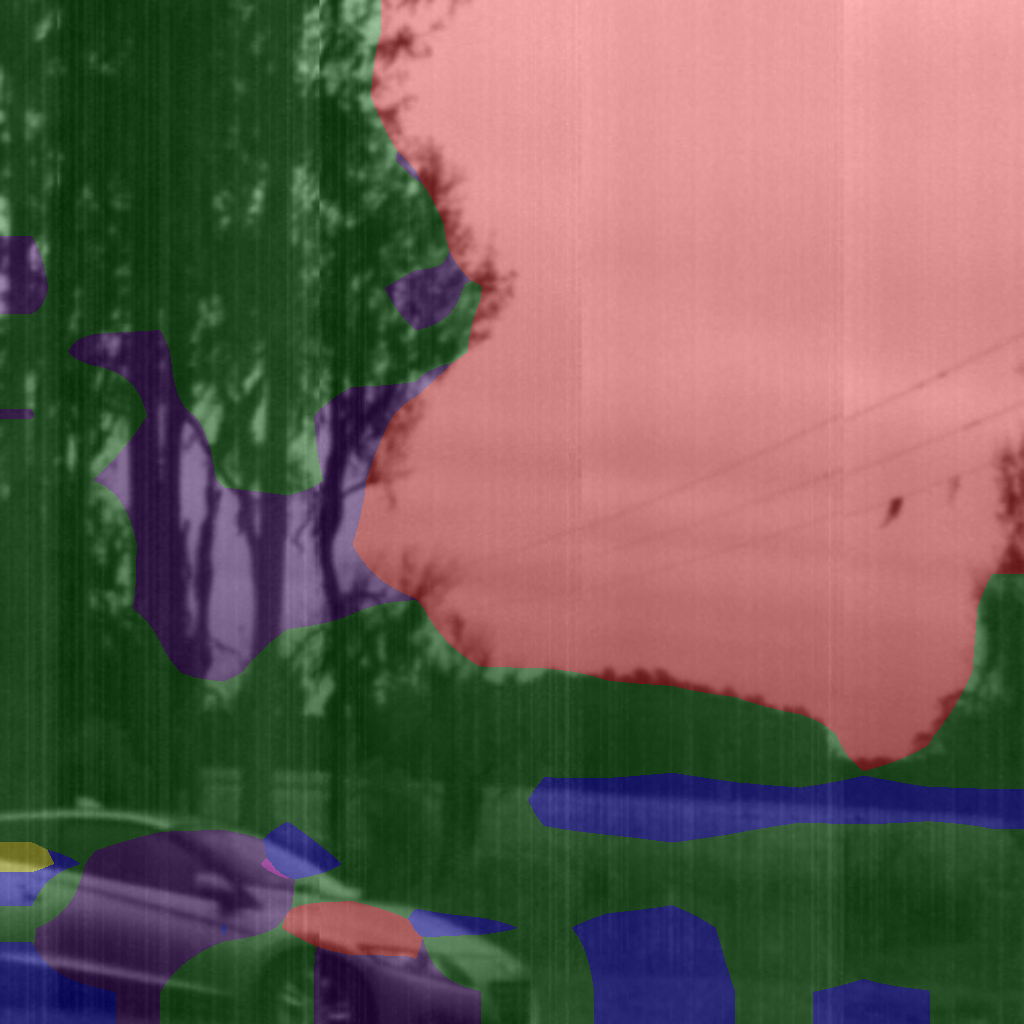}}\imghspace\setcounter{subfigure}{0}

\subfloat[Input image]{\includegraphics[width=0.16\linewidth]{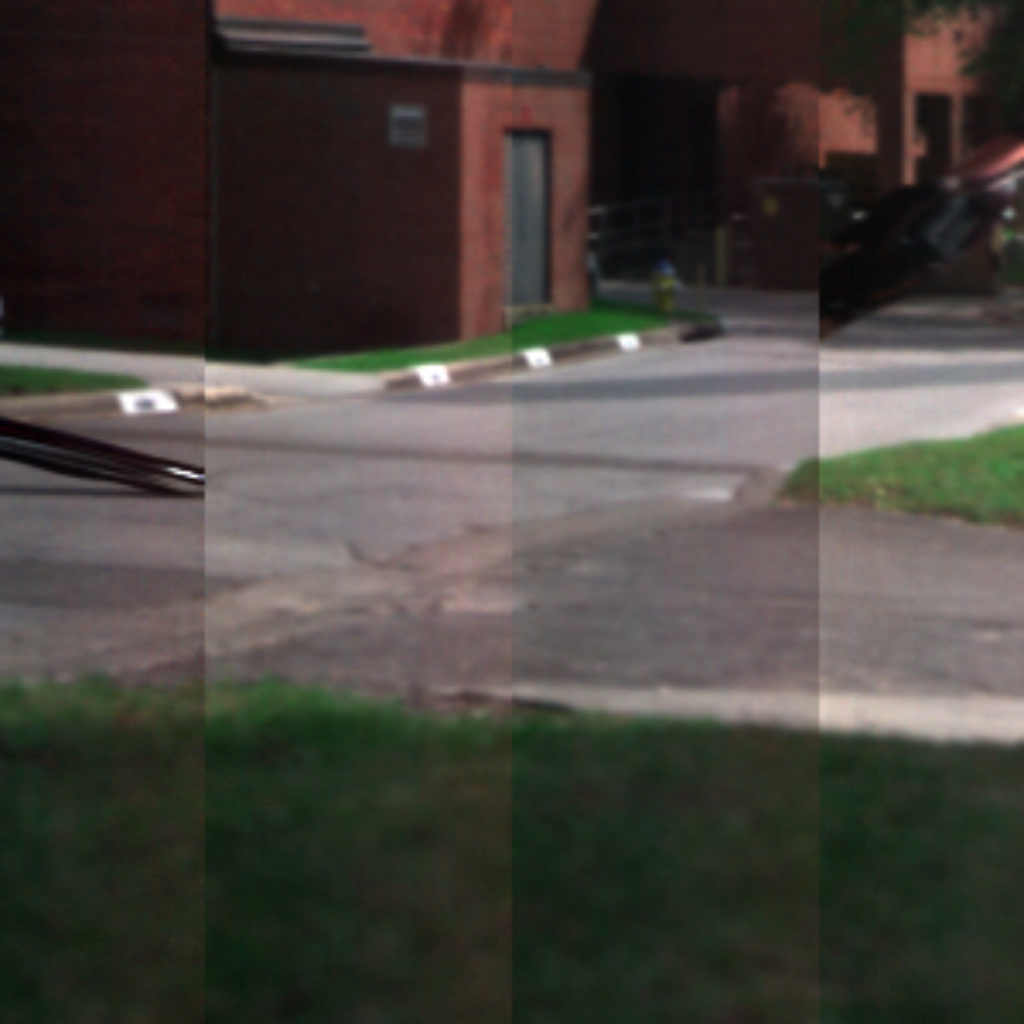}}\imghspace
\subfloat[Sailency Map]{\includegraphics[width=0.16\linewidth]{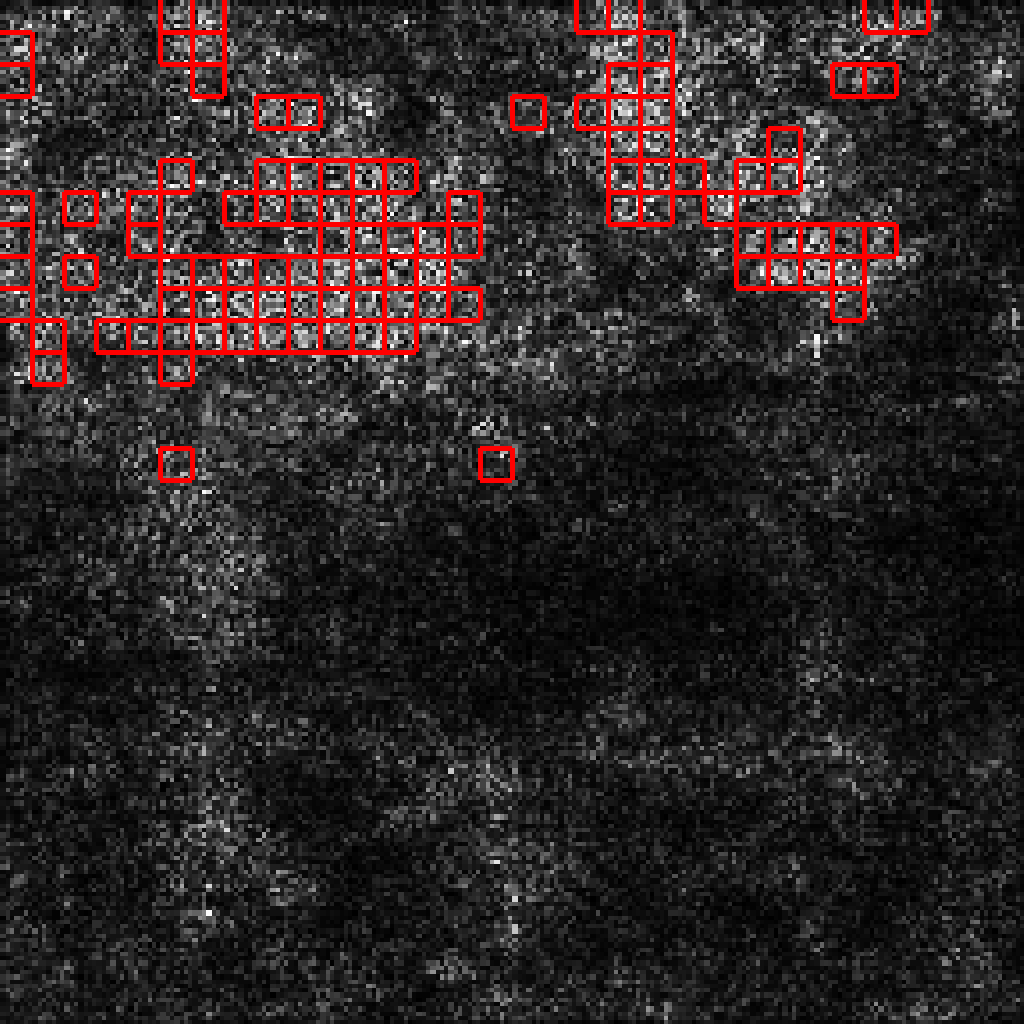}}\imghspace
\subfloat[Segmentations]{\includegraphics[width=0.16\linewidth]{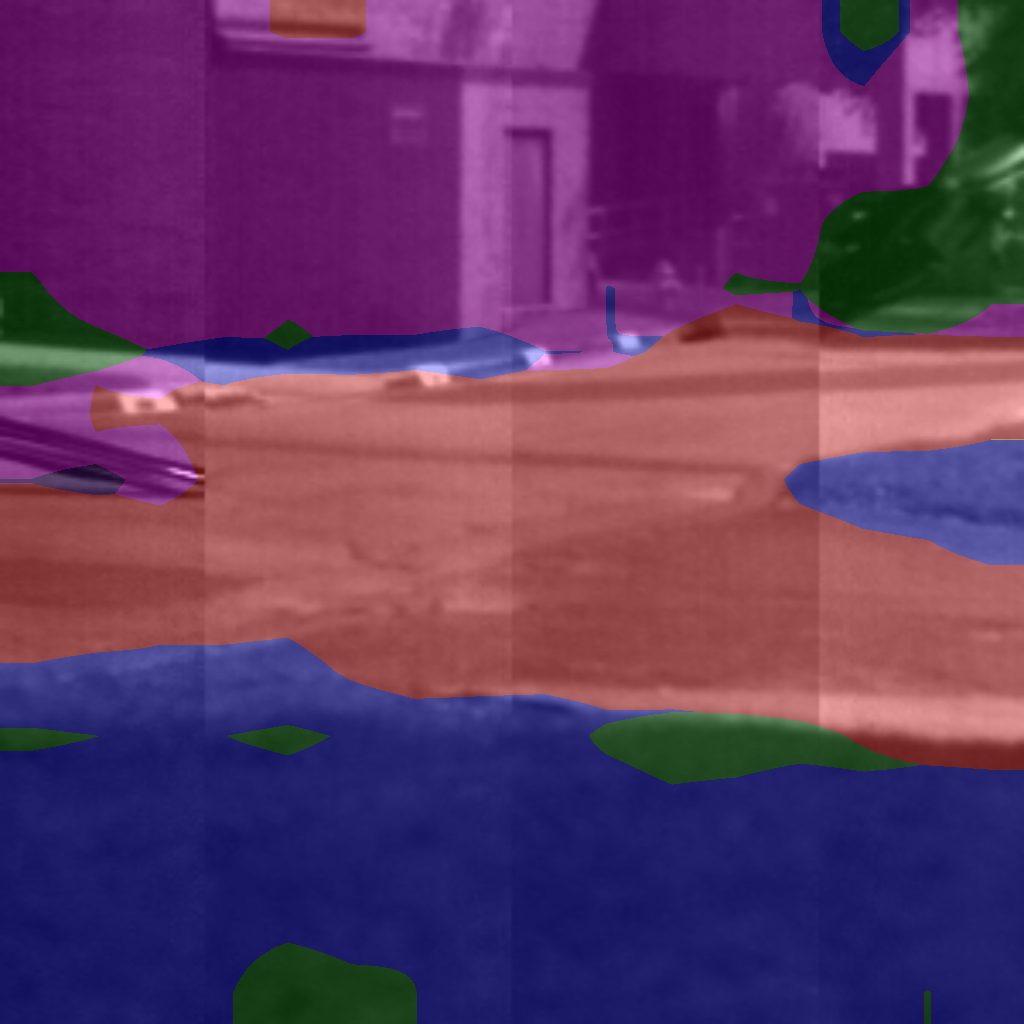}}\hspace{.01\linewidth}\setcounter{subfigure}{0}
\subfloat[Input image]{\includegraphics[width=0.16\linewidth]{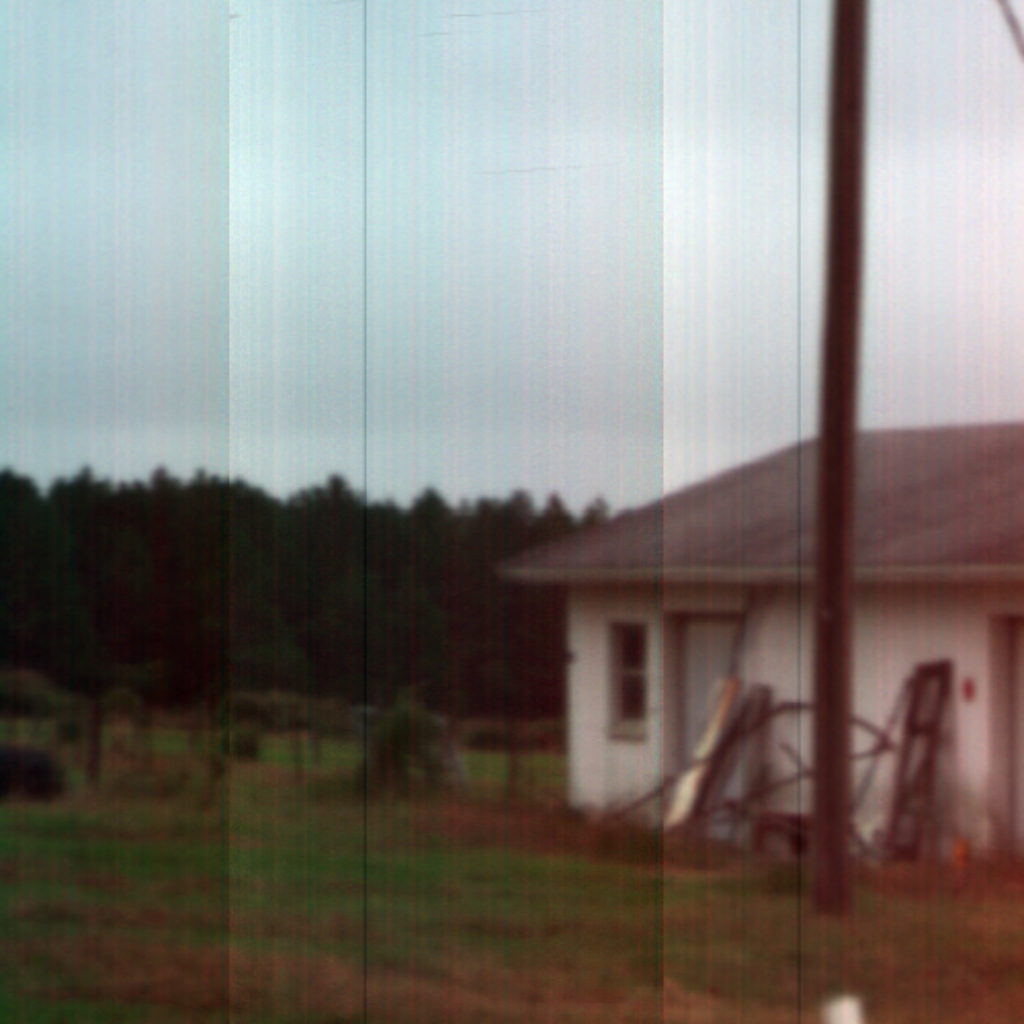}}\imghspace
\subfloat[Sailency Map]{\includegraphics[width=0.16\linewidth]{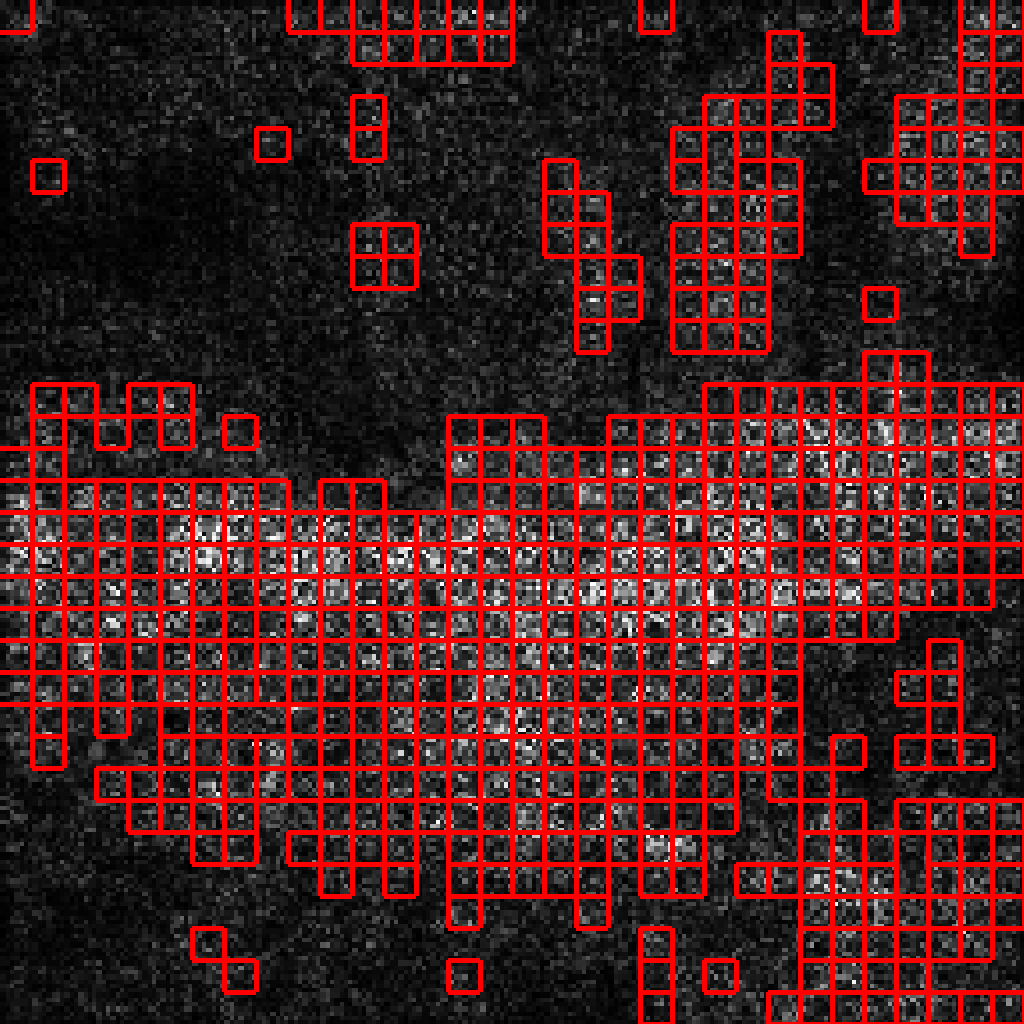}}\imghspace
\subfloat[Segmentations]{\includegraphics[width=0.16\linewidth]{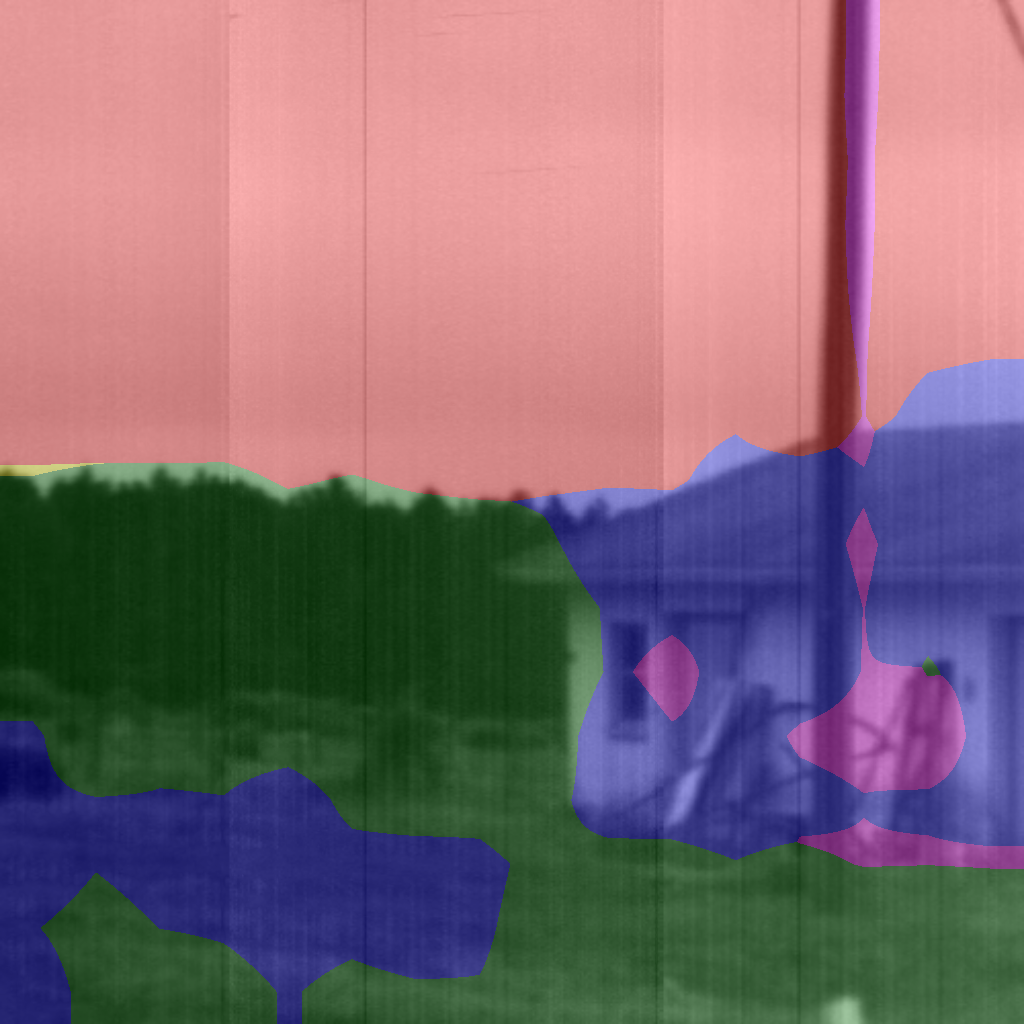}}\imghspace

\caption{Partially simulated results. Segmentations were done on a RGB-NIR, 4 channel image.}
\label{fig:real results}
\end{figure*}

\subsubsection{Attention Mechanisms}
To select the wandering patches we used three different selection criteria, edge detection, a static saliency map, and a trainable saliency map. The edge detection consisted of the image convolved with a sobel filter. The saliency maps were collected from the gradient data generated when an image is backpropagated through a network. The network we used is the pretrained VGG19 network. In the static map, the weights were not updated, and in the trainable, the weights were allowed to be updated based on the loss of the entire network. The maps generated from each of these three methods were then put through a size 32 and stride 32 max pool to score each patch position. The greatest positions were chosen to be the wandering patches.

\subsection{Efficiency Benefits}

Getting better results in a vision transformer comes by using a higher resolution image or using smaller patches in a greater number. In both of these cases, the amount of patches being used as input increases. For instance, if the initial patch inputs is $n$, halving the size of the patches, or doubling resolution of the image will result in $4n$ patch inputs. 

Our method trades off the benefits of high resolution patches without the need for every high-resolution patch as input. At the scales of HSI data, it is not practical to apply methods like ViT segmentation for high-resolution input. Instead, we suggest that deciding  which high-resolution patches the segmentation algorithm can use helps to address computational issues. We now explain our experiments to show that performance does not suffer with improvement in computational load.  

\subsection{Full Simulation Results}


The network was trained and evaluated on RANUS: RGB-NIR Urban Scene Dataset \cite{choe_ranus_2018} and HyKo hyperspectral drivability dataset \cite{hyko}.
Experiments were run with three different attention mechanisms and a variable number of wandering patches.
We consider the results with no wandering patches as a baseline comparison. \emph{Note, that rather than emphasizing novelty in segmentation, we are instead proposing a mechanism to maintain performance with less computation.} In other words, our approach is complementary to any other dataset or segmentation method. 

As we increase the number of wandering patches in our network, 
the results will approach that of the full high-resolution, 1024x1024 baseline results. All experiments were run with size 32 patches, and images of size 256 and 512 were used for the low-resolution images. 

\noindent \textbf{RANUS:} Table \ref{table:segmentation results} shows RANUS dataset results. Using the trainable saliency map, we achieved the best results of a mean IOU 25.61, which slightly exceeds the 1024 baseline 25.08, while using only $35\%$ of the patches in the 1024 baseline. 


\noindent \textbf{HyKo:} Table \ref{table:hyko_table} shows Hyko dataset results. This dataset consists of 15 color channels in the visible spectrum.

\section{Experiments with Camera Prototype}

In this section we use the modified ViT segmenter for hyperspectral images with the camera discussed earlier. We first show a result where the mirror only captures a high-resolution and low-resolution images, and patch selection is done in software. We then show a result where patch selection is done by mirror motion to the desired viewing direction, to capture the selected high-resolution patches.

\subsection{Partially Simulated Results}

In this experiment we capture data from both a high resolution image (10240x10240) and a low resolution image (2560x2560) of the same scene. After downsampling the images, we use the low resolution image as the input into the network, and the high resolution image as reference for the determined patches to create the segmentations. 

To get the high resolution scene, the mirror is moved at a constant rate. To get the lower resolution scene, the mirror is able to move faster, and samples are gathered every 4 mirror steps, speeding up the image acquisition. 

Fig \ref{fig:real results} shows these results. For this image we used the static attention mechanism with 100 patches (TA 100). The image was taken of a brick building approximately 100ft away from the camera. Our network is able to correctly identify the contiguous parts of the scene such as the building and sky.

\section{Limitations and Conclusion}

In this paper we have shown how a push-broom hyperspectral camera, integrated with a swiveling mirror, can enable HSI data with adaptive resolution. We proposed a variant of a multispectral transformer-based segmentation algorithm that was designed in concert with the camera, to enable segmentation with reduced computational load. 

Our work has the following limitations:

\begin{enumerate}
    \item We have only showed results on 15 color channels, whereas both our camera and our algorithm can take hundreds of bands. This is due to the limitation of hyperspectral semantic segmentation datasets, which we hope to correct in the future. 
    \item Our current setup only distributes resolution, and so, correctly does not do any ``zoom". Adding a focus tunable lens could fix this, but would open up issues related to dispersion across the different spectral bands.
\end{enumerate}

In conclusion, we have designed a new prototype and algorithm for efficient hyperspectral segmentation. We hope that these will impact areas where such cameras are widely used, such as agriculture and remote sensing.
{
    \small
    \bibliographystyle{IEEEtran}
    \bibliography{references}
}


\end{document}